\documentclass[10pt,journal,compsoc]{IEEEtran}
\ifCLASSOPTIONcompsoc
  \usepackage[nocompress]{cite}
\else
  \usepackage{cite}
\fi
\usepackage{amsmath,amssymb,amsfonts}
\usepackage{algorithm}
\usepackage{algorithmic}
\usepackage{graphicx}
\usepackage{textcomp}
\usepackage{xcolor}
\usepackage{bm}
\usepackage{threeparttable}
\usepackage{makecell}
\makeatletter
\newcommand{\removelatexerror}{\let\@latex@error\@gobble}
\makeatother

\newcommand{\eat}[1]{}

\ifCLASSINFOpdf
\else
\fi
\begin{document}
%
\title{A Survey on Curriculum Learning}
%
%
%
%

\author{Xin~Wang,~\IEEEmembership{Member,~IEEE},
        Yudong~Chen,
        and~Wenwu~Zhu,~\IEEEmembership{Fellow,~IEEE}
\IEEEcompsocitemizethanks{\IEEEcompsocthanksitem Xin Wang, Yudong Chen, Wenwu Zhu are with the Department
of Computer Science and Technology, Tsinghua University, Beijing, China.\protect\\
E-mail: xin\_wang@tsinghua.edu.cn, cyd18@mails.tsinghua.edu.cn, wwzhu@tsinghua.edu.cn. Corresponding Author: Wenwu Zhu.
This work is supported by the National Key Research and Development Program of China (No.2020AAA0106300, 2020AAA0107800,  2018AAA0102000).
}}

%
%


\markboth{IEEE TRANSACTIONS ON PATTERN ANALYSIS AND MACHINE INTELLIGENCE,~Vol.~, No.~}%
{Wang \MakeLowercase{\textit{et al.}}: A Survey on Curriculum Learning}
%



\IEEEtitleabstractindextext{%
\begin{abstract}
Curriculum learning (CL) is a training strategy that trains a machine learning model from easier data to harder data, which imitates the meaningful learning order in human curricula. As an easy-to-use plug-in, the CL strategy has demonstrated its power in improving the generalization capacity and convergence rate of various models in a wide range of scenarios such as computer vision and natural language processing etc. In this survey article, we comprehensively review CL from various aspects including motivations, definitions, theories, and applications. We discuss works on curriculum learning within a general CL framework, elaborating on how to design a manually predefined curriculum or an automatic curriculum. In particular, we summarize existing CL designs based on the general framework of {\it Difficulty Measurer $+$ Training Scheduler} and further categorize the methodologies for automatic CL into four groups, i.e., Self-paced Learning, Transfer Teacher, RL Teacher, and Other Automatic CL. We also analyze principles to select different CL designs that may benefit practical applications. Finally, we present our insights on the relationships connecting CL and other machine learning concepts including transfer learning, meta-learning, continual learning and active learning, etc., then point out challenges in CL as well as potential future research directions deserving further investigations.
\end{abstract}

}

\maketitle

\IEEEdisplaynontitleabstractindextext

%
\IEEEpeerreviewmaketitle

\IEEEraisesectionheading{\section{Introduction}\label{sec:1-introduction}}
Human learning has inspired various algorithm designs throughout the development of machine learning. As an outstanding feature of human learning, curriculum, or learning in a meaningful order, has been exploited and transferred to machine learning, which forms the subdiscipline named \emph{curriculum learning (CL)}. In essence, human education is highly organized as curricula, by ``starting small'' and gradually presenting more complex concepts. For example, to learn calculus at college, a student should first learn basic arithmetic at primary school, abstract function at middle school, and then derived function at high school. However, in traditional machine learning algorithms, all the training examples are randomly presented to the model, ignoring the various complexities of data samples and the learning status of the current model. Therefore, an intuitive question is: ``\emph{could the curriculum-like training strategy ever benefit machine learning?}'' According to the extensive experiments from early work~\cite{bengio2009curriculum,kumar2010self,zaremba2014learning} to recent efforts~\cite{platanios2019competence,graves2017automated,fan2018learning,hacohen2019power} in various applications of machine learning, we may summarize the answer as: ``yes, but not always.'' As we will demonstrate in this survey, the power of introducing curriculum into machine learning depends on how we design the curriculum for specific applications and datasets.


\begin{figure}[htbp]
\vspace{-2mm}
\centerline{\includegraphics[width=0.9\linewidth]{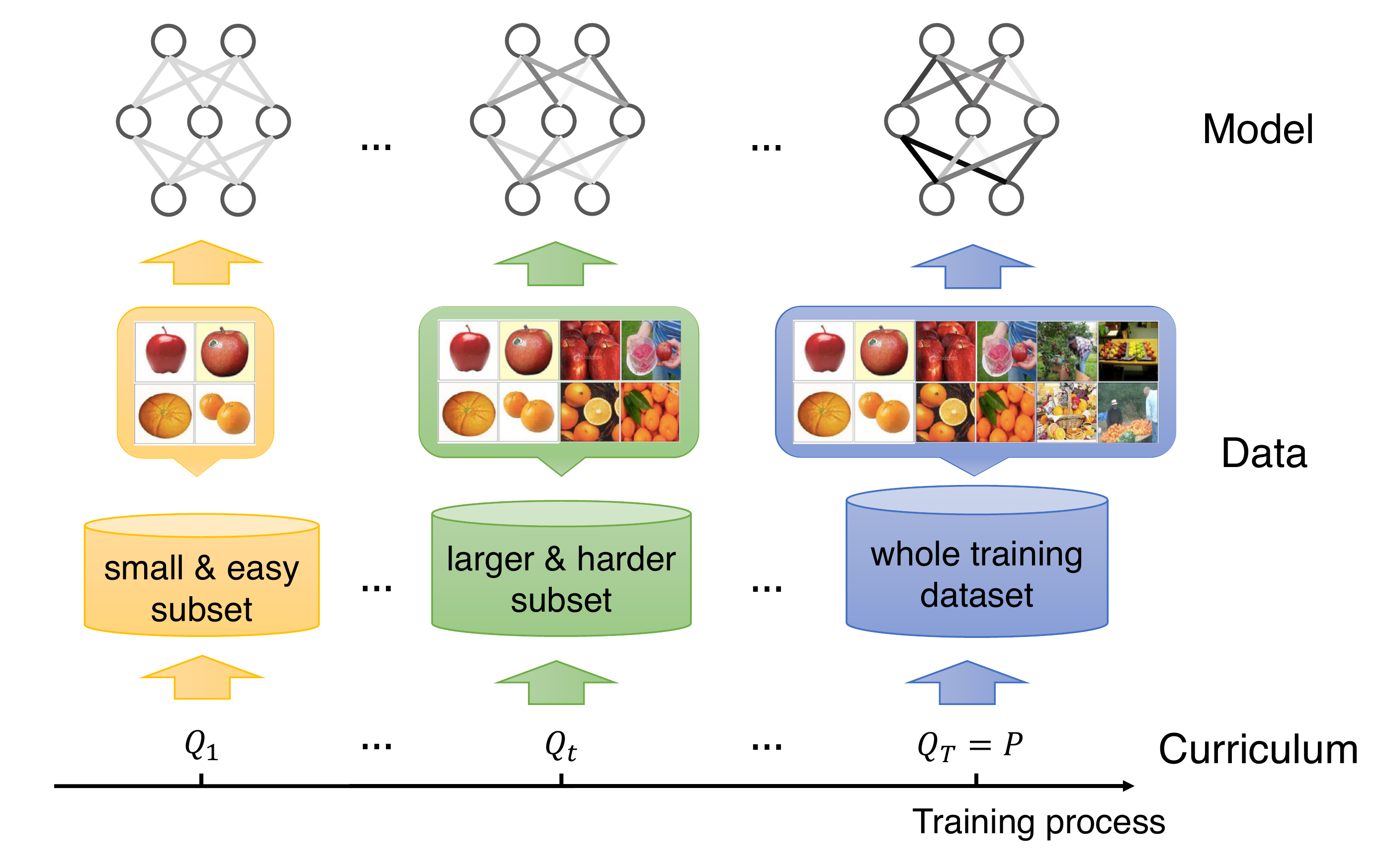}}
\vspace{-2mm}
\caption{Illustration of the Curriculum Learning (CL) concept (The fruit images are from~\cite{soviany2020image}). CL is a training strategy for machine learning that trains from easier data to harder data, imitating human curricula. Specifically, CL initially trains the model on a small and easy subset. With the progress of the training, CL gradually introduces more harder examples into the subset, and finally trains the model on the whole training dataset. This CL strategy can improve both model performance and convergence rate, compared with direct training on the whole training dataset. $Q_t$ here stands for a reweighting of the training data distribution $P$ at the $t$-th training epoch (See details in Sec.~\ref{sec:2-definition}).}
\label{fig:illustration_of_CL_concept}
\vspace{-2mm}
\end{figure}

The original concept of CL is first proposed by Bengio et al.~\cite{bengio2009curriculum}. In short, curriculum learning means ``training from easier data to harder data''. More specifically, the basic idea is to ``start small''~\cite{elman1993learning}, train the machine learning model with easier data subsets (or easier subtasks), and then gradually increase the difficulty level of data (or subtasks) until the whole training dataset (or the target task(s)). An illustration of CL is demonstrated in Fig.~\ref{fig:illustration_of_CL_concept}, where we take the image classification task as an example. Initially, CL trains the model on a small subset of ``easy'' images, i.e., the images of apples and oranges are clear, typical, and easily recognizable. With the progress of model training, CL adds more ``harder'' images (i.e., harder to recognize) to the current subset, which is akin to the increasing difficulty of learning materials in human curricula. Finally, CL leverages the whole training dataset for training. 


\begin{figure*}[htbp]
\vspace{-2mm}
\centerline{\includegraphics[width=0.75\linewidth]{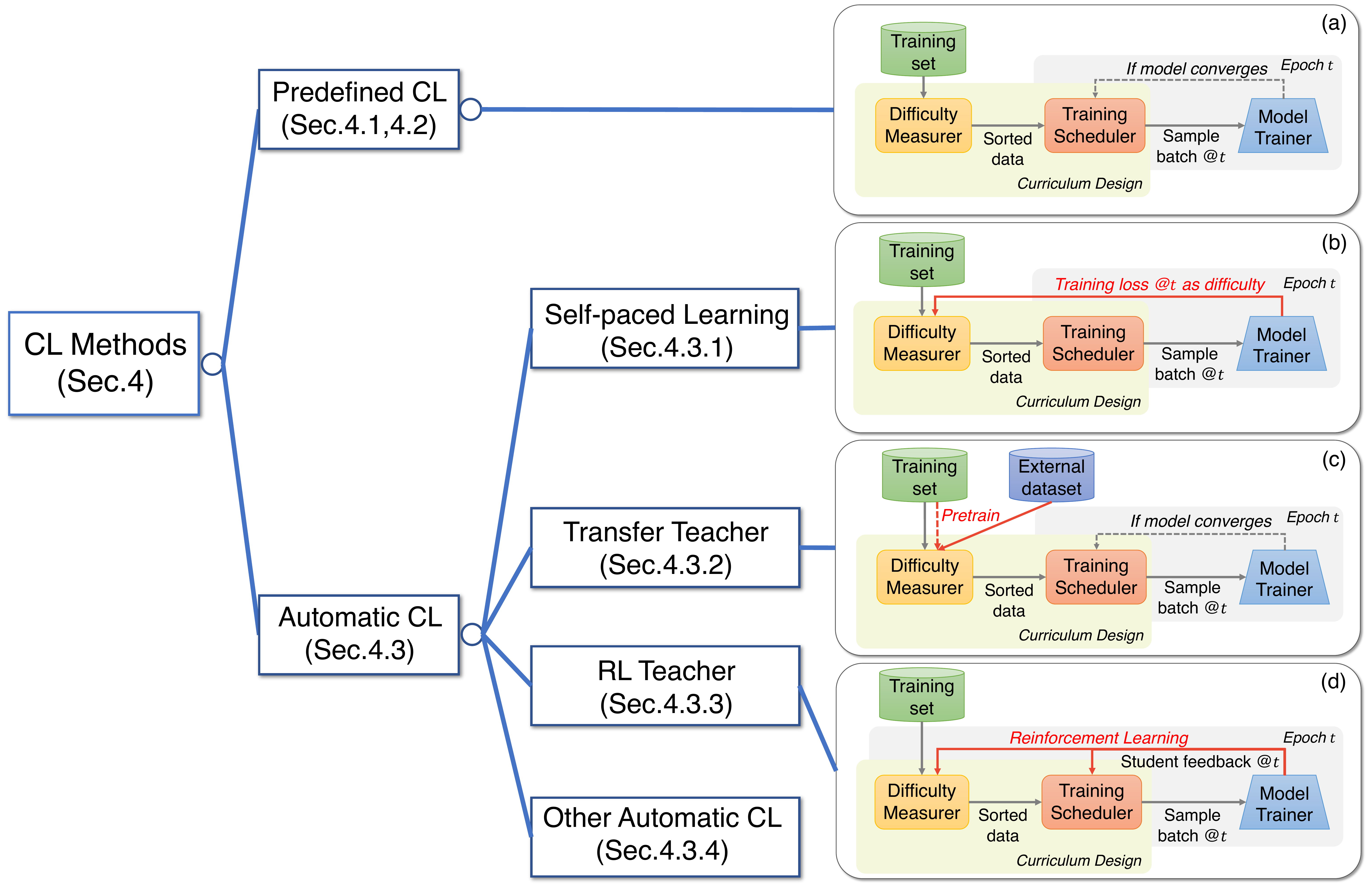}}
\vspace{-3mm}
\caption{A categorization of CL methods and the corresponding illustrations. We divide the existing methods into predefined CL and automatic CL, the latter of which including Self-paced Learning, Transfer Teacher, RL Teacher and Other Automatic CL. As shown in the illustrations, most CL methods comply with the general framework of {\it Difficulty Measurer $+$ Training Scheduler} in Sec.~\ref{sec:4-CL-design-a-general-framework}.}
\label{fig:CL_methods}
\end{figure*}

As the idea of CL serves as a general training strategy beyond specific machine learning tasks, scholars have been exploiting its power in considerably wide application scopes, including supervised learning tasks within computer vision (CV)~\cite{jiang2014easy,guo2018curriculumnet}, natural language processing (NLP)~\cite{platanios2019competence,tay2019simple}, healthcare prediction~\cite{el2020student}, etc., various reinforcement learning (RL) tasks~\cite{florensa2017reverse,narvekar2017autonomous,ren2018self} as well as other applications such as graph learning~\cite{qu2018curriculum, Gong2019MultiModalCL} and neural architecture search (NAS)~\cite{guo2020breaking}.
The advantages of applying CL training strategies to miscellaneous real-world scenarios can be mainly summarized as \emph{improving the model performance on target tasks} and \emph{accelerating the training process}, which cover the two most significant requirements in major machine learning research. 
For example, in~\cite{platanios2019competence}, CL helps the neural machine translation model reduce training time by up to 70\% and improves the performance by up to 2.2 BLEU points, compared to plain training without curricula. In~\cite{jiang2014self}, CL brings a relative 45.8\% MAP boost from normal batch training with an obvious faster convergence in multimedia event detection task. In~\cite{florensa2017reverse}, CL enables the RL agents to solve hard goal-oriented problems that they cannot solve without curricula. 
Apart from the above two main advantages, CL is also easy-to-use, since it is a flexible plug-and-play submodule independent of the original training algorithms in most CL literature. 
However, to the best of our knowledge, little effort has been made to systematically summarize the methodologies and applications of CL. 

In this paper, we fill this gap by comprehensively reviewing CL and summarizing its methodologies. To be more specific, we hope to provide the readers with an overall picture of CL, which includes comprehensible and elaborate answers to the following questions: 
(i) What is the definition of CL (Sec.~\ref{sec:2-definition})? 
(ii) Why is CL effective, and why should researchers use CL (Sec.~\ref{sec:3-analysis-and-application-scenes})?
(iii) How to design a curriculum (Sec.~\ref{sec:4-CL-design-a-general-framework})?
We conclude the paper with a comparison of ``easier first'' and ``harder first'' training strategies and discussion on the relationship between CL and other machine learning concepts in Sec.~\ref{sec:5-discussions}. We also summarize several open questions and future directions for CL to inspire future researchers in Sec.~\ref{sec:6-future-directions}.


\vspace{-2mm}
\section{Definition of CL}
\label{sec:2-definition}

\textbf{History context.} Empirical evidence supporting the meaningfulness of taking curricula in human and animal learning has been early provided in behavior and cognitive science literature. Skinner~\cite{skinner1958reinforcement,peterson2004day} provides the earliest behavior evidence on the importance of \textit{shaping}, i.e., another name for CL in animal training context. Cognitive evidence is then provided in human size constancy learning~\cite{turkewitz1982limitations} and language learning~\cite{newport1990maturational}.
The idea of introducing a curriculum into the training strategy of machine learning algorithms can be traced back to Selfridge et al.'s work~\cite{selfridge1985training}. The authors proposed to train a cart pole controller, a classic problem in robotics, first on long and light poles and then gradually on shorter and heavier poles. Later related work~\cite{schmidhuber1991curious,sanger1994neural} in RL and robotics domains also discussed how to organize the presenting order of tasks from easy to hard. The first attempt of the curriculum-like idea on supervised learning is made by Elman~\cite{elman1993learning} in the NLP task of grammar learning with recurrent networks. The author highlighted the importance of ``starting small'': restricting the range of data exposed to neural networks during initial training. This strategy is also revisited in~\cite{rohde1999language} and~\cite{krueger2009flexible}, the latter of which provides evidence for faster convergence. 

Based on all these previous works, the concept of CL was first proposed by Bengio et al.~\cite{bengio2009curriculum} with experiments on supervised visual and language learning tasks, exploring when and why a curriculum could benefit machine learning. The original definition of CL by Bengio et al.~\cite{bengio2009curriculum} is as follows.

\textbf{Definition 1: Original Curriculum Learning~\cite{bengio2009curriculum}}. A curriculum is a sequence of training criteria over $T$ training steps: $\mathcal{C}=\left\langle Q_1,\dots,Q_t,\dots,Q_T \right\rangle$. Each criterion $Q_t$ is a reweighting of the target training distribution $P(z)$:
\begin{equation}
\label{eq:original-CL-definition}
\footnotesize
Q_t(z) \propto W_t(z) P(z)\quad \forall \text{example}\ z \in \text{training set}\ D,
\end{equation}
such that the following three conditions are satisfied:
\vspace{-2mm}
\begin{itemize}
\item 1) The entropy of distributions gradually increases, i.e., $H(Q_t) < H(Q_{t+1})$. 
\item 2) The weight for any example increases, i.e., $W_t(z) \le W_{t+1}(z)\quad \forall z \in D$. 
\item 3)\ $Q_T(z) = P(z)$.
\end{itemize}
Curriculum learning is the training strategy that trains a machine learning model with a curriculum.

In Definition 1, Condition (1) means the diversity and information of the training set should gradually increase, i.e., the reweighting of examples in later steps increases the probability of sampling slightly more difficult examples.
Condition (2) means to gradually add (in binary or soft manner) more training examples, so the size of the training set increases.
Condition (3) means finally, the reweighting of all examples is uniform and we train on the target training set.


Most of the CL methods discussed in this paper (especially those in Sec.~\ref{subsec:4-2-manually-predefined-CL},~\ref{subsubsec:self-paced-learning}, and ~\ref{subsubsec:transfer-teacher}) meet Definition 1, 
illustrated in Fig~\ref{fig:illustration_of_CL_concept}. As shown in the figure, the CL strategy determines the training data subset of each training step, such that the size and overall difficulty of the subsets are gradually increasing throughout the training process.

Since the concept of CL was formally proposed, the academic community follows and further extends the definition of CL. 
Within the spirit of ``training from easier data (tasks) to harder data (tasks)'', i.e., fixing Condition (1) in Definition 1, Condition (2) and (3) can be relaxed to enable more flexible CL strategies. 
For example, in~\cite{zaremba2014learning,pentina2015curriculum,graves2017automated} of multi-task setting and most CL for RL settings~\cite{florensa2018automatic}, Condition (2) and (3) are relaxed since at each step the model is trained on only one task. However, the diversity or difficulty of the current task/goal gradually increases, which guides the model to boost the performance on the target task(s). 
The CL methods based on One-Pass scheduler~\cite{tudor2016hard,tay2019simple,wang2020curriculum} discussed in Sec.~\ref{subsubsec:predefined-training-scheduler} also relaxes Condition (2) and (3) as they train the model from easier subsets to harder subsets. 
Moreover, other works also extend Definition 1 by adding more conditions of data characteristics for different application purposes. For instance, Jiang et al.~\cite{jiang2014self} propose to train ``from easy \& diverse to hard'' to avoid overfitting to the same sample group in multi-group event detection tasks. Wang et al.~\cite{wang2019dynamic} train the model ``from easy \& imbalanced to hard \& balanced'' data to alleviate the severe class imbalance in human attribute analysis.

At a more abstract level, a curriculum can be seen as a sequence of (binary) \emph{instance selection}~\cite{olvera2010review} or (soft) \emph{example reweighting} along the training process to achieve faster convergence or better generalization, which is beyond the ``easy to hard'' or ``starting small'' principles. This perspective inspires the academic community to bring more connotations to CL definition with new methodologies, which can be summarized as follows.

\textbf{Definition 2: Data-level Generalized Curriculum Learning.} Discarding all the three conditions in Definition 1, a curriculum is a sequence of reweighting of target training distribution over $T$ training steps. Curriculum learning is the strategy that trains a model with such a curriculum.

Most CL methods in Sec.~\ref{subsubsec:RL-teacher} and Sec.~\ref{subsubsec:other-automatic-CL} could learn to automatically and dynamically select the most suitable examples or tasks (with adjustable loss weights) for each current training step and thus meet Definition 2. 
Interestingly, in some of the works, the best curriculum found by the algorithm is the opposite of traditional CL, i.e., ``hard to easy''~\cite{fan2018learning,wang2019dynamically} or ``starting big'' (from full dataset to informative subset)~\cite{wang2020learning,zhao2020reinforced,wang2019dynamically}. 
There is also a line of research named hard example mining (HEM)~\cite{shrivastava2016training,jin2018unsupervised} selecting the most difficult examples in each training batch. HEM actually falls in Definition 2 and is explored in some CL literature~\cite{jesson2017cased, Zhou2020CurriculumLB}.
A discussion on this seemingly paradoxical phenomenon will be made in Sec.~\ref{subsec:5-1-easier-first-vs-harder-first}.


To even further broaden the scope of CL, some scholars jump from data level to criteria level, to regard a curriculum as a sequence of \emph{training criteria} during the training process. This further generalizes the CL definition:

\textbf{Definition 3: Generalized Curriculum Learning.} Discarding the definition of $Q_t$ (Eq.~\ref{eq:original-CL-definition}) and its three conditions in Definition 1, a curriculum is a sequence of training criteria over $T$ training steps. Each criterion $Q_t$ includes the design for all the elements in training a machine learning model, e.g., data/tasks, model capacity, learning objective, etc. Curriculum learning is the strategy that trains a model with such a curriculum.

Examples for training criteria in Definition 3 include, but are not limited to, loss function~\cite{wu2018learning,saxena2019data}, supervision generation~\cite{Zhang2020SynthesizingSF,Han2019WeaklySupervisedLO}, model capacity~\cite{morerio2017curriculum,Sinha2020CurriculumBS,Karras2018ProgressiveGO}, input scheme~\cite{bengio2015scheduled}, and hypothesis space~\cite{guo2020breaking}. Note that the criteria in such a generalized curriculum in Definition 3 usually change progressively, analogous to the gradual curriculum in human education. For example, in Curriculum Dropout~\cite{morerio2017curriculum}, the algorithm gradually reduces the ratio of active units in dropout operation from 1 to a predefined $\theta_0 \in (0,1)$ to achieve adaptive regularization during training. 
In Curriculum NAS~\cite{guo2020breaking}, the algorithm starts from a small search space and gradually incorporates the learned knowledge to guide the search in larger spaces, which significantly improves the search efficiency and also finds better neural architectures. 
These works broaden the extension of CL and exploit the potentialities of the human curriculum idea for machine learning at a higher level, leaving room for imagination for future work.


\vspace{-3mm}
\section{Analysis on Effectiveness of CL and Suitable Application Scenes}
\label{sec:3-analysis-and-application-scenes}

Before applying CL to their studies, researchers might be curious about a fundamental question: why on earth does this human-curriculum-like training strategy work? 
To explain why CL could lead to generalization improvement and convergence speedup, scholars have provided hypotheses and proofs from different perspectives. Basically, existing analyses uncover the essence of CL from the perspectives of \emph{optimization problem} and \emph{data distribution}, based on which we can further summarize the two main motivations for applying CL: \emph{to guide} and \emph{to denoise}. 

\vspace{-2mm}
\subsection{Theoretical Analysis on CL}
\label{subsec:3-1-theoretical-analysis}

\begin{figure}[htbp]
\vspace{-2mm}
\centerline{\includegraphics[width=0.7\linewidth]{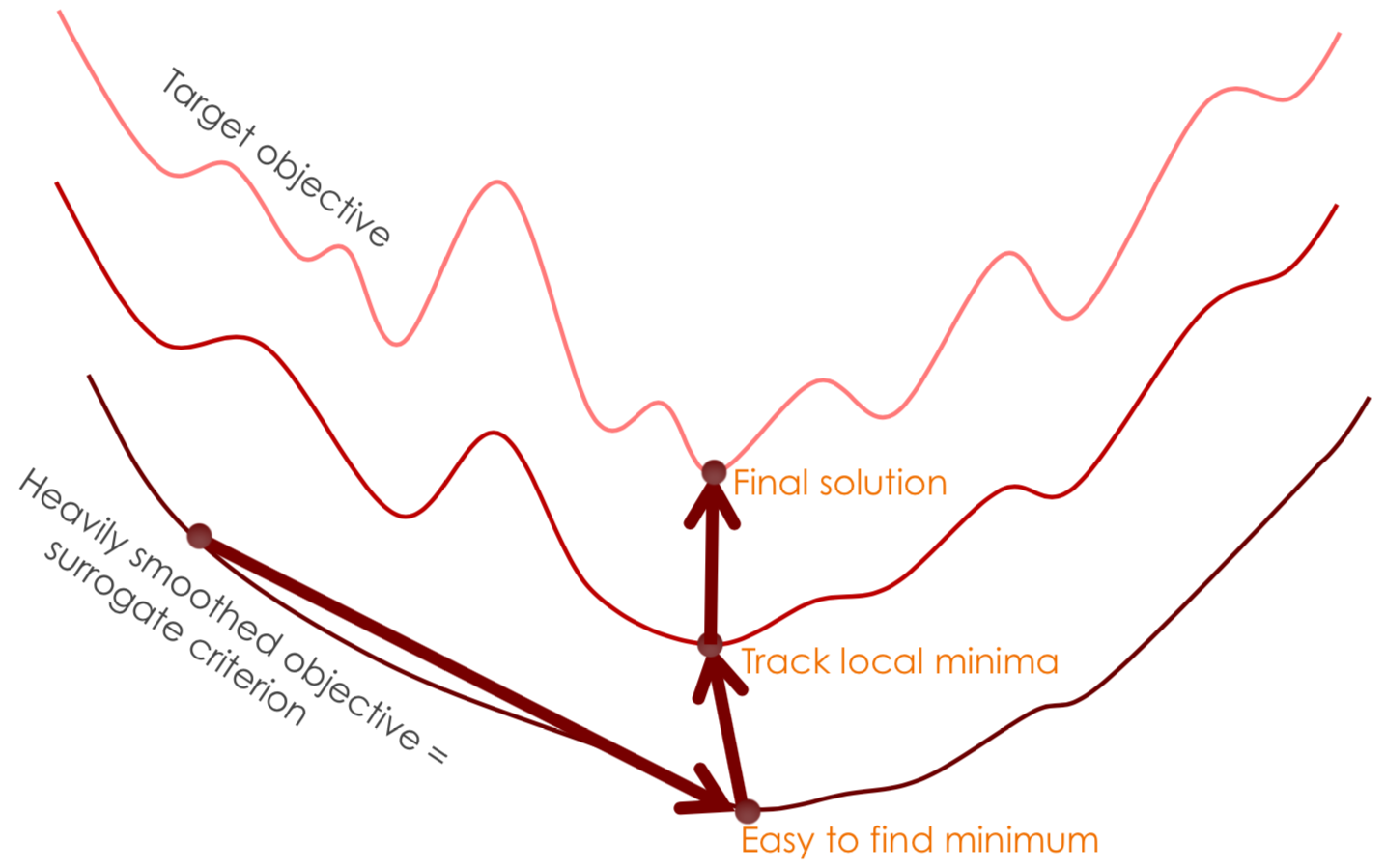}}
\vspace{-2mm}
\caption{Illustration of the continuation method from~\cite{bengio2014evolving}, which is the essence of the CL~\cite{bengio2009curriculum}. It starts from optimizing a heavily smoothed version of the objective, and gradually moves to the target objective. Tracking the local minima throughout the training guides the model towards better parameter space and makes it more generalizable.}
\label{fig:continuation_method}
\vspace{-2mm}
\end{figure}

To begin with, from the perspective of \textbf{optimization problem}, Bengio et al.~\cite{bengio2009curriculum} initially point out that CL can be seen as a particular \emph{continuation method}. Intuitively, continuation methods~\cite{allgower2012numerical} are optimization strategies for non-convex criteria which first optimize a smoother (and also easier) version of the problem to reveal the ``global picture'', and then gradually consider less smoothing versions, until the target objective of interest. 
This strategy also shares the same spirit with \emph{simulated annealing}. As illustrated in Fig~\ref{fig:continuation_method}, continuation methods provide a sequence of optimization objectives, starting with a heavily smoothed objective for which it is easy to find a global minimum, and tracking the local minima throughout the training. In this way, continuation methods \emph{guide} the training towards better regions in parameter space, i.e., as shown in Fig~\ref{fig:continuation_method}, the local minima learned from easier objectives have better generalization ability and are more likely to approximate global minima. Moreover, from the view of transfer learning, this continuation strategy can also be regarded as a sequence of unsupervised pre-training~\cite{bengio2009curriculum}: training on the preceding objectives could act as a pre-training process which both helps optimization and provides regularization on succeeding objectives. 

Additionally, recent studies provide more theoretical evidence for the convergence speedup in CL from the optimization perspective. Weinshall et al.~\cite{weinshall2018curriculum} prove a theorem 

On the other hand, researchers also analyze the CL mechanism from the perspective of \textbf{data distribution}. 
In the era of deep learning, large-scale data sources are required for training, which are collected and annotated by company users, the web, and crowd-sourcing systems. This big data collection brings noisy data that is less cognizable or wrongly annotated. 
In the CL setting, the noisy data corresponds to harder examples in the datasets while the cleaner data form the easier part. Since CL strategy encourages training more on the easier data, an intuitive hypothesis is that CL learner wastes less time with the harder and noisy examples to achieve faster training~\cite{bengio2009curriculum}. This hypothesis reveals the \emph{denoising} efficacy of CL on noisy data.

\begin{figure}[htbp]
\vspace{-2mm}
\centerline{\includegraphics[width=0.8\linewidth]{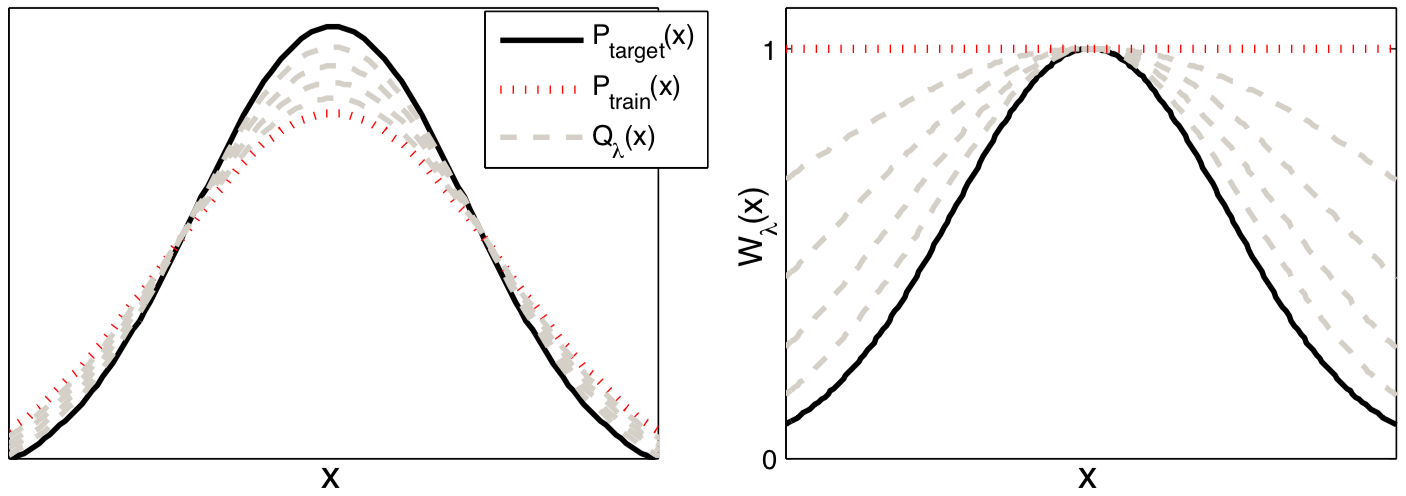}}
\vspace{-3mm}
\caption{Illustration of the CL from the data distribution perspective~\cite{gong2016curriculum}. The left part demonstrates the data distribution shifts from the easy subset (the solid curve, which is assumed to approximate the testing distribution $P_{\text{target}}(x)$ well) to the full training set $P_{\text{train}}(x)$ (the red dashed curve). The right part shows the corresponding weighting scheme to enable this distribution shift. The center peak of curves refers to the high-confidence clean data, while the tails refer to the noisy data in the distributions. As shown in the left part, $P_{\text{target}}(x)$ is cleaner than $P_{\text{train}}(x)$.}
\label{fig:denoise}
\vspace{-3mm}
\end{figure}

To have a closer look at this denoising mechanism, Gong et al.~\cite{gong2016curriculum} provide a theory based on the assumption that there exists deviation between training and testing distributions caused by noisy/wrongly-annotated training data. Intuitively, training and target/testing distributions share a common high-confidence annotated region with large density, which corresponds to the easier examples in CL. Therefore, to start training from easier examples by CL strategy actually simulates learning from this high-confidence common region (as an approximation to the target distribution), which guides the learning towards the expected target while reduces the negative impacts from low-confidence noisy examples. This data distribution perspective of CL is illustrated in Fig~\ref{fig:denoise}. The common density peak (at the center of the x-axis) of training and target distributions $P_{\text{train}}(x)$ and $P_{\text{target}}(x)$ in the left part refers to the common high-confidence area, while the heavy tail of $P_{\text{train}}(x)$ demonstrates the relatively more noisy data in training distribution. The right part illustrates the sequence of weight functions in CL, which initially assigns small values to the noisy tails and much larger values in the common easy area, and gradually moves to equal weights for all examples. Based on the above analysis, the authors formulate $P_{\text{target}}(x)$ as the weighted expression of $P_{\text{train}}(x)$. A follow-up theory clarifies that CL essentially minimizes an upper bound of the expected risk under target distribution, and this bound shows that we could approach the task of minimizing the expected risk on $P_{\text{target}}(x)$ by taking the core idea of CL: gradually taking relatively easy examples according to the curriculum and minimizing the empirical risk on these examples. 

\subsection{Suitable Application Scenes of CL}
\label{subsec:3-2-application-scenes}

Based on the above analysis on why CL is effective, we can categorize the motivations for applying CL into two groups: \textbf{to guide}, regularizing the training towards better regions in parameter space (with steeper gradients) as from the perspective of the optimization problem, and \textbf{to denoise}, focusing on high-confidence easier area to alleviate the interference of noisy data as from the perspective of data distribution. Not surprisingly, most of the existing application scenes of CL can be classified into these two groups, as demonstrated in Table~\ref{table:suitable_application_scenes}.

\begin{table*}[htbp]
\caption{Suitable Application Scenes of CL.}
\scriptsize
\vspace{-5mm}
\begin{center}
\begin{tabular}{p{1.5cm}p{3cm}p{4cm}p{7cm}}
\hline
\textbf{Motivation} & \textbf{Effect} & \textbf{Scene} & \textbf{Examples} \\
\hline
To guide & make training possible / better and faster & the target task is hard or has a different distribution & sparse reward RL, multi-task learning, GAN training, NAS; domain adaption, imbalanced classification \\
\hline
To denoise & make training faster, more robust and generalizable & tasks with noisy, uneven quality, heterogeneous data (often large-scale, cheaply collected) & weakly-supervised or unsupervised learning, NLP tasks (neural machine translation, natural language understanding, etc.)  \\
\hline
\end{tabular}
\label{table:suitable_application_scenes}
\end{center}
\vspace{-5mm}
\end{table*}

The application scenes based on the ``to guide'' motivation often involve difficult target tasks where direct training on these tasks results in poor performance or slow convergence. CL strategies are adopted to guide the training from easier tasks or smoother versions of objectives to the target tasks. For instance, in sparse-reward RL, direct training on the final tasks rarely gets any positive rewards, which hinders agent learning. Therefore, researchers propose to take the CL strategy and manually~\cite{matiisen2019teacher} or automatically~\cite{florensa2017reverse} design a sequence of auxiliary (sub)tasks/goals from easy to hard to guide the training. 
In multi-task learning, learning all the tasks simultaneously or in random order often leads to unsatisfactory performance. To yield performance gains, CL strategies are adopted to automatically choose the easier tasks which are more related to the previous one~\cite{pentina2015curriculum} or can bring more learning progress to the model training~\cite{graves2017automated,matiisen2019teacher}. 
Other examples include CL for training GANs~\cite{Karras2018ProgressiveGO,soviany2020image,ghasedi2019balanced} and NAS~\cite{guo2020breaking}.



Besides, the ``to guide'' application scenes also include the tasks where the target distribution is quite different from the training distribution, and a good curriculum helps to guide the training for adaption to the target distribution. A representative scene is domain adaption, which aims at improving prediction performance on unlabeled target domain data by knowledge transfer from richly annotated source domain data with a distribution drift. Recent studies~\cite{shu2019transferable,zhang2019curriculum} propose to train from more in-domain data (similar to target domain) to less in-domain data, guiding the model to adapt to the target domain while adequately exploiting the source domain data. Note that CL for domain adaption is also related to the ``to denoise'' motivation, if we regard the less in-domain data as a kind of noisy data. Another example is imbalanced classification problems, where the training distribution on different classes is extremely imbalanced. 
Different studies adopt various curricula either beginning from balanced subset to more imbalanced full dataset~\cite{jesson2017cased} or from easy and imbalanced subset to harder and more balanced subset~\cite{wang2019dynamic} to improve the generalization capacity of the classifier. 

On the other hand, the application scenes based on the ``to denoise'' motivation often have a noisy or heterogeneous training dataset, and CL strategies could help denoise, making the training faster, more robust, and more generalizable. A popular application of CL with this motivation is neural machine translation (NMT), whose dataset is highly heterogeneous in quality, difficulty, and noise~\cite{kumar2019reinforcement}. This is because the translation of a sentence could be long and short with different vocabulary and grammar structures, and different annotators always provide translations of different qualities 
Moreover, the training of NMT models (e.g., RNNs) is often time-consuming. 
Therefore, CL is naturally suitable for NMT tasks to denoise during training and to achieve both performance boost and faster convergence. Similarly, CL is also adopted in other NLP tasks with noisy or heterogeneous data, including natural language understanding~\cite{xu2020curriculum}, relation extraction~\cite{huang2019self}, reading comprehension~\cite{tay2019simple}, etc. Moreover, CL is also effective in weakly-supervised CV tasks~\cite{liang2016learning,guo2018curriculumnet}. 

From the perspective of supervision in training, CL can help supervised, weakly-supervised, and unsupervised learning by guiding or denoising.
Specifically, CL helps supervised setting mainly by guiding when (i) the task is hard~\cite{florensa2017reverse,pentina2015curriculum}, (ii) parts of the training data are difficult to learn~\cite{bengio2009curriculum,jiang2014self}, (iii) the target distribution heavily shifts from training distribution~\cite{shu2019transferable,wang2019dynamic}. 
Weakly-supervised setting includes three typical types~\cite{zhou2018brief}, all of which are enhanced by CL denoising. 
(i) For inaccurate supervision, i.e., the training set is noisy and usually collected from the web, CL helps to denoise, enabling the model to focus on a cleaner subset to avoid bad local minimum~\cite{guo2018curriculumnet,shu2019transferable,liang2016learning,platanios2019competence}. 
(ii) For incomplete supervision, i.e., the semi-supervised setting where some training data are unlabeled, CL helps to distinguish the easier (more confident) unlabeled examples and add them to the training set earlier or with higher weights, denoising the pseudo labels with low confidence (harder unlabeled data)~\cite{gong2016multi,Gong2019MultiModalCL,Zhang2020FewCostSO,tudor2016hard}.
(iii) For inexact supervision, i.e., only coarse-grained labels are given, CL helps to gradually integrate confident fine-grained pseudo labels into training while denoising the noisy ones, usually under a multi-instance learning framework~\cite{zhang2015self,zhang2019leveraging,Han2019WeaklySupervisedLO,tang2018attention}. 
Finally, CL can also help unsupervised setting, e.g., clustering~\cite{ghasedi2019balanced,xu2015multi}, feature selection~\cite{zheng2020unsupervised}, domain adaption~\cite{choi2019pseudo}, etc. The mechanism in most work is similar to the semi-supervised setting, i.e., denoising the noisy pseudo labels~\cite{ghasedi2019balanced,zheng2020unsupervised,choi2019pseudo}. The function to guide is also explored in~\cite{xu2015multi}. With carefully designed CL,~\cite{Zhang2020SynthesizingSF} even learns deep saliency network without human annotation by progressively synthesizing supervision masks.


\vspace{-3mm}
\section{CL Design: a General Framework}
\label{sec:4-CL-design-a-general-framework}

Since we have understood why CL is effective and why researchers apply CL to different scenes, a natural and important question should be: how to design an appropriate curriculum for a specific learning task? In this section, we provide a general framework of ``Difficulty Measurer + Training Scheduler'' (Sec.~\ref{subsec:4-1-general-framework}), which unifies most of CL methodologies. 
Based on this framework, we categorize the existing CL methods into predefined CL (Sec.~\ref{subsec:4-2-manually-predefined-CL}) and automatic CL (Sec. ~\ref{subsec:4-3-automatic-CL}) and introduce the representative designs in each category. Fig.~\ref{fig:CL_methods} illustrates the typology of CL methods introduced in this section.

\subsection{The General Framework of Difficulty Measurer + Training Scheduler}
\label{subsec:4-1-general-framework}

Recall that the core definition of CL (Definition 1) lies in the strategy of ``training from easier data to harder data''. In essence, to design such a curriculum, we need to decide two things: 1) What kind of training data is supposed to be easier than other data? 2) When should we present more harder data for training, and how much more? Issue 1) can be abstracted to a \textbf{Difficulty Measurer}, which decides the relative ``easiness'' of each data example. 
Issue 2) can be abstracted to a \textbf{Training Scheduler}, which decides the sequence of data subsets throughout the training process based on the judgment from the Difficulty Measurer. 


Therefore, a general framework for curriculum design consists of these two core components: Difficulty Measurer + Training Scheduler, which is illustrated in Fig~\ref{fig:CL_methods}(a). To begin with, all the training examples are sorted by the Difficulty Measurer from the easiest to the hardest and passed to the Training Scheduler. Then, at each training epoch $t$, the Training Scheduler samples a batch of training data from the relatively easier examples and sends it to Model Trainer for training. With the progress of training epochs, Training Scheduler will decide when to sample from more harder data, (usually) until uniform sampling from the whole training set. This schedule sometimes also depends on the training loss feedback from the Model Trainer (the dashed arrow in Fig~\ref{fig:CL_methods}(a)), e.g., Training Scheduler presenting more harder data when the current model converges. 
Note that in~\cite{hacohen2019power}, the authors conclude the two core components as \emph{scoring function} and \emph{pacing function}, which share the same spirit with Difficulty Measurer and Training Scheduler, respectively, while the latter names adopted in this paper are chosen to be more abstract and clearer.

Let us take the experiment in Fig~\ref{fig:illustration_of_CL_concept} as an instantiation example for our CL framework. Difficulty Measurer is the human annotations deciding that some fruit images in the dataset are easier than other images based on recognizability and complexity. Training Scheduler can be, for example, a linear scheduler (see Sec.~\ref{subsubsec:predefined-training-scheduler}) that starts with 40\% of easiest examples in each class, and increases this proportion by 5\% each epoch until 100\%. 
In this way, an effective curriculum is designed by instantiating the general CL framework according to the specific image classification task.

According to our framework, we could also clarify the scopes of predefined CL and automatic CL in the next two sections. Specifically, when both the Difficulty Measurer and Training Scheduler are designed by human prior knowledge with no data-driven algorithms involved, we call the CL method \textbf{predefined CL}. If any (or both) of the two components are learned by data-driven models or algorithms, then we denote the CL method as \textbf{automatic CL}. 

\subsection{Predefined CL}
\label{subsec:4-2-manually-predefined-CL}

In this section, we discuss the common types of manually predefined Difficulty Measurers (Sec.~\ref{subsubsec:predefined-difficulty-measurer}) and Training Schedulers (Sec.~\ref{subsubsec:predefined-training-scheduler}) under our CL framework, 
and conclude the main limitations of predefined CL (Sec.~\ref{subsubsec:limitation-of-predefined-CL}).

\subsubsection{Common Types of Predefined Difficulty Measurer}
\label{subsubsec:predefined-difficulty-measurer}

Researchers have manually designed various Difficulty Measurers mainly based on the data characteristics of specific tasks. We summarize common types of Difficulty Measurers in Table~\ref{table:difficulty-measurer}. Most of the predefined Difficulty Measurers are designed for image and text data in various CV and NLP scenarios, while other data types include audio data, programs, tabular data, etc. 
Interestingly, we find that except for some domain knowledge-based measurement (marked as ``Domain''), most of the predefined Difficulty Measurers are designed from the angles of complexity, diversity, and noise estimation, which are separate but also correlated. 

\begin{table}[htbp]
\caption{Common types of predefined Difficulty Measurer. The ``+'' in $\propto$Easy means the higher the measured value, the easier the data example, and the ``-'' has the opposite meaning.}
\vspace{-5mm}
\scriptsize
\begin{threeparttable}
\begin{center}
\begin{tabular}{p{3.9cm}p{1.5cm}p{1.3cm}p{0.5cm}}
\hline
\textbf{Difficulty Measurer}\tnote{*} & \textbf{Angle} & \textbf{Data Type} & $\propto$\textbf{Easy} \\
\hline

Sentence length~\cite{spitkovsky2010baby,platanios2019competence} & Complexity & Text & - \\
Number of objects~\cite{wei2016stc} & Complexity & Images & - \\
\# conj.~\cite{kocmi2017curriculum}, \#phrases~\cite{tsvetkov2016learning}  & Complexity & Text & - \\
Parse tree depth~\cite{tsvetkov2016learning} & Complexity & Text & - \\
Nesting of operations~\cite{zaremba2014learning} & Complexity & Programs & - \\

Shape variability~\cite{bengio2009curriculum} & Diversity & Images & - \\
Word rarity~\cite{platanios2019competence,kocmi2017curriculum} & Diversity & Text & - \\
POS entropy~\cite{tsvetkov2016learning} & Diversity & Text & - \\
Mahalanobis distance~\cite{el2020student} & Diversity & Tabular & - \\

Cluster density~\cite{guo2018curriculumnet,choi2019pseudo} & Noise & Images & + \\
Data source~\cite{chen2015webly}  & Noise &  Images & / \\
SNR / SND ~\cite{braun2017curriculum,ranjan2017curriculum} & Noise & Audio & - \\

Grammaticality~\cite{liu2018curriculum} & Domain & Text & + \\ 
Prototypicality~\cite{tsvetkov2016learning} & Domain & Text & + \\ 
Medical based~\cite{jimenez2019medical} & Domain & X-ray film & / \\
Retrieval based~\cite{penha2019curriculum,ferro2018continuation} & Domain & Retrieval & / \\

Intensity~\cite{gui2017curriculum} / Severity~\cite{tang2018attention} & Intensity & Images & + \\
Image difficulty score~\cite{tudor2016hard,soviany2020image} & Annotation & Images & - \\
Norm of word vector~\cite{liu2020norm} & Multiple & Text & - \\

\hline
\end{tabular}
\label{table:difficulty-measurer}
\begin{tablenotes}
    \footnotesize
    \item[*] Abbreviations: POS = Part Of Speech, SNR = Signal to Noise Ratio, SND = Signal to Noise Distortion, Domain = Domain knowledge, \# conj. = number of coordinating conjunctions.
\end{tablenotes}
\end{center}
\end{threeparttable}

\vspace{-3mm}
\end{table}

Firstly, \emph{complexity} stands for the structural complexity of a particular data example, such that examples with higher complexity have more dimensions and are thus harder to be captured by models. For instance, sentence length, the most popular Difficulty Measurer in NLP tasks~\cite{spitkovsky2010baby, tay2019simple, platanios2019competence},  
intuitively expresses the complexity of a sentence/paragraph. Therefore, longer sentences are often supposed as harder training data. Other examples include the number of objects in images in the task of semantic segmentation~\cite{wei2016stc}; the number of coordinating conjunctions (e.g., ``and'', ``or'')~\cite{kocmi2017curriculum} or phrases (e.g., prepositional phrases)~\cite{tsvetkov2016learning}; the parse tree depth~\cite{tsvetkov2016learning} 
that measures the sentence complexity in the view of grammar; and the nesting of operations in program text~\cite{zaremba2014learning} that measures the complexity of the instruction set in program execution tasks. 

Secondly, the angle of \emph{diversity} here stands for the distributional diversity of a group of data (e.g., regular or irregular shapes~\cite{bengio2009curriculum}) or the elements (e.g., words) of a data point (e.g., sentence). A larger value of diversity means the data is more various, including more (rare) types/styles of data or elements, and is thus more difficult for model learning. For example, a sentence with more rare words is usually considered harder to learning~\cite{platanios2019competence}. A popular measure of diversity is information entropy, which is exploited both in text data as the Part-Of-Speech (POS) entropy~\cite{tsvetkov2016learning} and in tabular data as the Mahalanobis distance of feature vectors~\cite{el2020student}. Intuitively, both high complexity and high diversity bring more degrees of freedom to the data, which needs a model with larger capacity and bigger effort of training. 

Larger diversity sometimes also makes the data noisier. Therefore, another angle is \emph{noise estimation}, which estimates the noise level of data examples and defines cleaner data as easier. A quite intuitive method is taken in~\cite{chen2015webly} to judge the noise level by the source of image data on the web: images retrieved by a search engine like Google are supposed to be cleaner, and images posted on photo-sharing website like Flickr are more realistic and noisier. In~\cite{guo2018curriculumnet}, the authors map images to vectors by CNNs and suppose that cleaner images often appear similar, and thus have larger values of local density. Therefore, examples with lower local density are supposed to be noisier and harder to predict. Moreover, the Signal to Noise Ratio/Distortion (SNR/SND)~\cite{braun2017curriculum,ranjan2017curriculum} is widely adopted to estimate the noise in audio data.

Other interesting Difficulty Measurers include signal intensity~\cite{gui2017curriculum,tang2018attention} and human-annotation-based Image Difficulty Scores~\cite{tudor2016hard,soviany2020image}, both designed for image data. Signal intensity can be regarded as a measurement for the informativeness of data features. For example, in the task of facial expression recognition~\cite{gui2017curriculum}, more intense/exaggerated faces are supposed to be easier data than poker faces. In the task of thoracic disease diagnosis~\cite{tang2018attention}, more severe symptoms provide more information and are easier to recognize. Moreover, Image Difficulty Score~\cite{tudor2016hard} is proposed to measure the difficulty of an image by collecting the response times of human annotators in the following protocol: (i) ask the annotator ``Is there an \{object class\} (e.g., elephant) in the next image?'' and (ii) record the time spent by the annotator to answer ``Yes'' or ``No'' and use this response time to estimate Image Difficulty Score: intuitively, longer response time corresponds to harder image example. After collecting the annotation, the authors train a regression model to map the CNN features of new images to the difficulty score. 


\subsubsection{Common Types of Predefined Training Scheduler}
\label{subsubsec:predefined-training-scheduler}

While predefined Difficulty Measurers vary among different data types and tasks, the existing predefined Training Schedulers are usually data/task agnostic, i.e., the majority of CL literature in various scenarios leverages similar types of Training Schedulers. Generally, Training Schedulers can be divided into \emph{discrete} and \emph{continuous} schedulers. 
The difference is: discrete schedulers adjust the training data subset after every fixed number ($> 1$) of epochs or convergence on the current data subset, while continuous schedulers adjust the training data subset at every epoch.

\emph{Discrete schedulers} are widely adopted owing to their simplicity and effectiveness. The most popular discrete scheduler is named as \emph{Baby Step}~\cite{bengio2009curriculum,spitkovsky2010baby} (Algorithm 1), which first distributes the sorted data into buckets (or shards/bins) 
from easy to hard and starts training with the easiest bucket. After a fixed number of training epochs or convergence, 
the next bucket is merged into the training subset. Finally, after all the buckets are merged and used, the whole training process either stops or further continues several extra epochs. Note that at each epoch, the scheduler usually shuffles both the current buckets and the data in each bucket and then sample mini-batches for training (instead of using all data at once). 
\vspace{-2mm}

\begin{figure}[ht]
\vspace{-5mm}
  \renewcommand{\algorithmicrequire}{\textbf{Input:}}
  \renewcommand{\algorithmicensure}{\textbf{Output:}}
  \removelatexerror
  \begin{algorithm}[H]
  \footnotesize
    \caption{The Baby Step Training Scheduler~\cite{cirik2016visualizing}.}
    \begin{algorithmic}[1]
      \REQUIRE $\mathcal{D}$: training dataset; $\mathcal{C}$: the Difficulty Measurer; 
      \ENSURE $\bm{M^*}$: the optimal model.
      \STATE $\mathcal{D}' = \text{sort}(\mathcal{D}, \mathcal{C})$;
      \STATE $\{\mathcal{D}^1, \mathcal{D}^2, \cdots, \mathcal{D}^k\} = \mathcal{D}'$ where $\mathcal{C}(d_a) < \mathcal{C}(d_b),\ d_a \in \mathcal{D}^i, d_b \in \mathcal{D}^j, \forall i < j$;
      \STATE $\mathcal{D}^{train} = \emptyset$;
      \FOR {$s = 1 \cdots k$}
      	\STATE $\mathcal{D}^{train} = \mathcal{D}^{train} \cup \mathcal{D}^s$;
      	\WHILE {not converged for $p$ epochs}
      		\STATE $\text{train}(M, \mathcal{D}^{train})$;
      	\ENDWHILE
      \ENDFOR
    \end{algorithmic}
  \end{algorithm}
\vspace{-5mm}
\end{figure}

Another discrete scheduler called \emph{One-Pass}~\cite{bengio2009curriculum} takes a similar strategy of data bucketing from easy to hard and starting training from the easiest bucket. However, 
when updating, 
One-Pass scheduler discards the current bucket and switches to the next harder bucket. One-Pass is less used than Baby Step in CL literature (see \cite{zaremba2014learning,tudor2016hard,tay2019simple,wang2020curriculum} for One-Pass examples), probably due to the lower performance in many tasks. Intuitive reasons might include: 1) The complexity/diversity of the training data is gradually increasing in Baby Step scheduler, which helps improve generalization capacity; 2) The One-Pass scheduler is like training on a sequence of independent tasks as in continual learning~\cite{delange2021continual}, which faces the problem of catastrophic forgetting even though the early tasks are easier. 
The two schedulers are compared on LSTMs in~\cite{cirik2016visualizing}.

Other discrete schedulers are also based on data bucketing but take different sampling strategies. For example, in~\cite{kocmi2017curriculum}, the authors modify the Baby Step to unevenly divide the examples into buckets such that easier buckets have more data examples, which is natural to reach in the case of machine translation corpora. Then they sample examples without replacement from the easiest bucket only until there remain the same number of examples as in the second most easy bucket. Afterward, they uniformly sample from the first two buckets until the size is the same as that of the third bucket. In an empirical study of CL on NMT tasks~\cite{zhang2018empirical}, the authors also test other extensions of Baby Step, including 1) ``boost'': to copy the hardest bucket for further training; 2) ``reduce and add-back'': to gradually remove one easiest bucket from training set once all buckets have been used, and then add them back and repeat the removing until convergence; 3) ``no-shuffle'': to discard inter-bucket shuffling and always present from easier to harder buckets to the model. A conclusion is, including Baby Step, no single scheduler consistently outperforms others.




\emph{Continuous schedulers}, on the other hand, can be mostly regarded as a function $\lambda(t)$ to map training epoch number $t$ to a scalar $\lambda \in (0, 1]$, which means $\lambda$ proportion of easiest training examples are available at the $t$-th epoch. According to the Definition 1 in Sec.~\ref{sec:2-definition}, this function $\lambda(t)$ must be monotone and non-decreasing, starting at $\lambda(0) > 0$ and ending at $\lambda(T) = 1$ 
This function is also called \emph{pacing function}~\cite{hacohen2019power} or \emph{competence function}~\cite{platanios2019competence} in literature. 

Existing $\lambda(t)$ functions are various, while researchers could design new functions for their specific tasks. The most intuitive function is the \emph{linear function}, where $\lambda_0$ is the initial proportion of available easiest examples, and $T_{\text{grow}}$ denotes the epoch when the function reaches $1$ for the first time.

\vspace{-2mm}
\begin{equation}
    \scriptsize
	\lambda_{\text{linear}}(t) = \min \left(1, \lambda_0 + \frac{1-\lambda_0}{ T_{\text{grow}}} \cdot t \right)
\end{equation}

\emph{Root function} is later proposed in~\cite{platanios2019competence} according to the observation that in linear function, the newly added examples are less likely to be sampled as the training data subset grows in size. Therefore, to give the model sufficient time to learn the newly added examples, the authors reduce the number of newly added examples as training progresses by defining the rate of adding examples to be inversely proportional to the size of the current training subset: $\frac{d\lambda(t)}{dt} = \frac{P}{d\lambda(t)}$, where $P \ge 0$ is a constant. Then we get:
\begin{equation}
    \scriptsize
	\lambda_{\text{root}}(t) = \min \left(1, \sqrt{\frac{1-\lambda_0^2}{T_{\text{grow}}}\cdot t + \lambda_0^2} \right).
\end{equation}
\vspace{-2mm}

To make the curve even sharper, a more general form \emph{root-p function} is also considered as follows, where $p \ge 1$:
\begin{equation}
    \scriptsize
	\lambda_{\text{root-p}}(t) = \min \left(1, \sqrt{\frac{1-\lambda_0^p}{T_{\text{grow}}}\cdot t + \lambda_0^p} \right).
\end{equation}
Interestingly, in~\cite{penha2019curriculum} the authors oppositely propose to give easier examples more training time, by taking the following \emph{geometric progression function}:
\begin{equation}
    \scriptsize
	\lambda_{\text{geom}}(t) = \min \left(1, 2^{\left(\frac{\log_2 1 - \log_2 \lambda_0}{T_{\text{grow}}}\cdot t + \log_2 \lambda_0\right)} \right).
\end{equation}
\vspace{-2mm}

\begin{figure}[htbp]
\vspace{-3mm}
\centerline{\includegraphics[width=0.7\linewidth,height=0.5\linewidth]{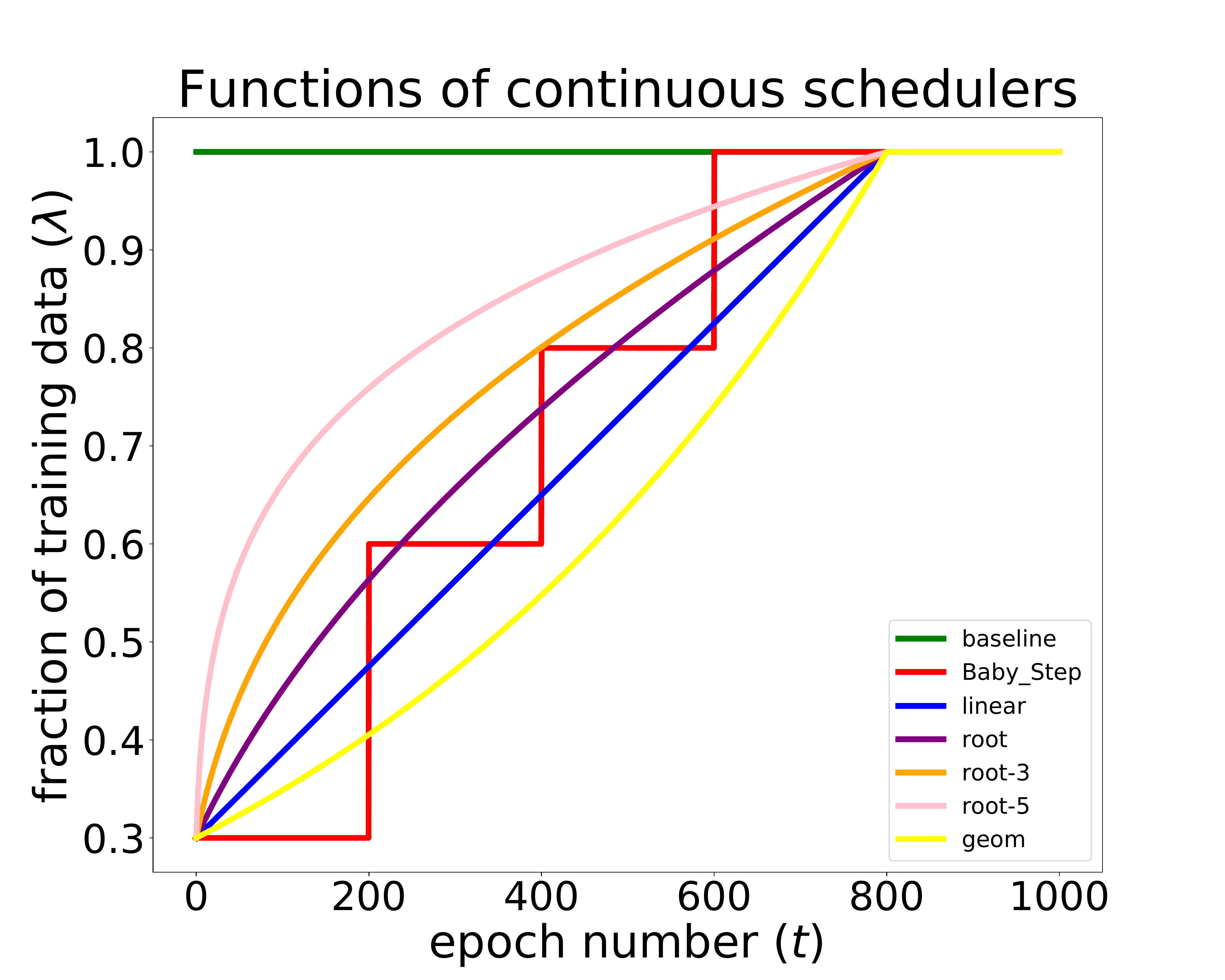}}
\vspace{-3mm}
\caption{Visualization of common continuous schedulers. The horizontal axis $t$ stands for the training epoch number, and the vertical axis $\lambda$ is the corresponding proportion of the easiest training data subset. Baseline is without curriculum and involves the whole training set from the beginning. The Baby Step scheduler is also visualized for comparison.}
\label{fig:continuous_schedulers}
\vspace{-3mm}
\end{figure}

The above continuous scheduler functions are illustrated in Figure~\ref{fig:continuous_schedulers}. Note that training without CL (``baseline'') and Baby Step are also regarded as special cases of continuous schedulers. The experiments in~\cite{platanios2019competence} and~\cite{penha2019curriculum} on 
NLP tasks show that the root-$p$ function ($p \ge 2$) is the most beneficial predefined Training Scheduler for CL, though the relative improvement to other schedulers is not drastic. 

Moreover, there is also a special group of continuous schedulers which do not follow the original definition of CL but perform as a sequence of data selection as in Definition 2. We name these schedulers as \emph{distribution shift}, which start training on an initial distribution and gradually move to a target distribution. For example, in~\cite{liu2018curriculum}, all the examples are divided into 2 groups: Common (lower quality and simpler) and Target (higher quality and more complex). The sampling weights are initially distributed on the Common and gradually shifted to the Target. In~\cite{jesson2017cased}, to alleviate extreme data imbalance in the lung nodule detection task, the scheduler starts sampling purely from images with nodules to learn to represent nodules, and then gradually decreases the proportion of examples with nodules until the extremely imbalanced data distribution (rare nodule). 

\vspace{-1mm}
\subsubsection{Limitations of predefined CL}
\label{subsubsec:limitation-of-predefined-CL}

Despite the simplicity and effectiveness of the predefined CL, there are some essential limitations as follows. 
(i) It is difficult to find the most suitable combination of Difficulty Measurer and Training Scheduler for a specific task and its dataset. There are no existing methodologies for selecting Difficulty Measurer and Training Scheduler other than exhaustive trials.
(ii) Both the predefined Difficulty Measurers and Training Schedulers stay fixed during the training process, which is not flexible enough and to some extent ignores the feedback of the current model.
(iii) Expert domain knowledge is often necessary for designing a predefined Difficulty Measurer. Moreover, when the dimension of example features is large, it is hard to predefine a computable Difficulty Measurer even by an expert.
(iv) Easy examples for humans are not always easy for models, since the decision boundaries of models and humans are basically different~\cite{yuan2019adversarial}.
(v) The best hyperparameters\footnote{The hyperparameters include $\lambda_0$, $T_{\text{grow}}$ and $p$ (in root-$p$ function) in continuous schedulers, and the number of steps, the number of epochs in each step in Baby Step based schedulers.} of Training Scheduler are hard to find. Additionally, a basic problem in Baby Step scheduler is to decide the number of buckets and how to divide the buckets\footnote{Division by thresholds on difficulty scores makes it hard to assign each bucket with roughly the same number of examples, while division by size may result in fluctuations in difficulty within a bucket or not enough difference between different buckets~\cite{zhang2018empirical}. An alternative is the Jenks Natural Breaks classification algorithm, as adopted in~\cite{zhang2018empirical}.}. 
(vi) 
The performance of various predefined Training Schedulers is sensitive to the initial learning rate (in NMT task)~\cite{zhang2018empirical}.

These limitations of predefined CL have prevented CL from being explored in more various applications. A natural and critical question is: how can we design more automatic Difficulty Measurers and Training Schedulers, which are more data- and model-driven instead of human-driven, more dynamically adaptive to the current training, and need fewer or even no hyperparameters to fine-tune? 

\vspace{-2mm}
\subsection{Automatic CL}
\label{subsec:4-3-automatic-CL}

In this section, we take a further step on the curriculum design by introducing automatic CL methods to break through the limits of predefined CL. A general comparison of predefined CL and automatic CL is presented in Table~\ref{table:comparison-of-predefined-and-automatic-CL}.

\begin{table}[htbp]
\vspace{-3mm}
\caption{Predefined CL v.s. automatic CL.}
\vspace{-4mm}
\begin{center}
\begin{tabular}{p{1.4cm}p{2.8cm}p{3.3cm}}
\hline
\textbf{Issues} & \textbf{Predefined CL} & \textbf{Automatic CL}  \\
\hline
Applicability & Need expert domain knowledge & General, domain agnostic \\
Difficulty Measurer & Human defined, fixed & Model decided, dynamic \\
Training Scheduler & Ignore model feedback, fixed & Consider model feedback, dynamic \\
\hline
\end{tabular}
\label{table:comparison-of-predefined-and-automatic-CL}
\end{center}
\vspace{-2mm}
\end{table}

We summarize the four major methodologies for automatic CL. 
In predefined CL, the \emph{teacher} designing the curriculum is a human expert, and the \emph{student} getting trained by the curriculum is the machine learning model. To reduce the need for human teachers, the four methodologies take different ideas, which can be intuitively summarized as follows. 
(i) \textbf{Self-Paced Learning (SPL)} methods let the student himself act as the teacher and measure the difficulty of training examples according to its losses on them. This strategy is analogous to the self-study of human students: one decides his/her own learning pace based on his/her current status.
(ii) \textbf{Transfer Teacher} methods invite a strong teacher model to act as the teacher and measure the difficulty of training examples according to the teacher's performance on them. The teacher model is pretrained and transfers its knowledge to measure example difficulty for student model training. 
(iii) \textbf{RL Teacher} methods adopt reinforcement learning (RL) models as the teacher to play dynamic data selection according to the feedback from the student. This strategy is the most ideal scene in human education, where the teacher and student improve together through benign interactions: the student makes the biggest progress based on the tailored learning materials selected by the teacher, while the teacher also effectively adjusts her teaching strategy to teach better.
(iv) \textbf{Other Automatic CL} methods include various automatic CL strategies except for the above-mentioned. The works take different optimization techniques to automatically find the best curriculum for model training, including Bayesian Optimization, meta-learning, hypernetworks, etc. Taking Definition 2 or 3, the curriculum in these methods often refers to a sequence of loss weights or even loss functions on data batches.

The comparison of these automatic CL methodologies is in Table~\ref{table:comparison-of-automatic-CL-methods}. Automatic CL is also broadly applied to Deep RL tasks, and we refer readers to the recent surveys~\cite{portelas2020automatic,narvekar2020curriculum} for further reading. The automatic CL methods discussed in this section are mostly designed for (weakly- or un-) supervised learning settings, though some of them are also shown to be effective for RL tasks~\cite{matiisen2019teacher,kim2018screenernet}.

\begin{table*}[htbp]
\vspace{-2mm}
\newcommand{\tabincell}[2]{\begin{tabular}{@{}#1@{}}#2\end{tabular}} 
\footnotesize
\caption{Comparison of the automatic CL methodologies, except ``Other Automatic CL''.}
\vspace{-4mm}
\begin{center}
\begin{tabular}{p{2.6cm}p{3.2cm}p{3.2cm}p{6.5cm}}
\hline
\textbf{Issues} & \textbf{Self-Paced Learning} & \textbf{Transfer Teacher} & \textbf{RL Teacher} \\
\hline
\textbf{Characteristic} & Student-driven difficulty & Teacher-driven difficulty & Teacher select data according to student feedback  \\
\textbf{Difficulty Measurer} & Automatic & Automatic & Automatic \\
\textbf{Training Scheduler} & Predefined & Predefined & Automatic \\
\textbf{Strength} & Efficient, robust & Reliable difficulty & Flexible \\
\textbf{Weakness} & Fixed strategy & Extra pretraining & Costly (Deep RL) \\
\textbf{CL Definition} & Definition 1 & Definition 1 & Definition 2 \\
\hline
\end{tabular}
\label{table:comparison-of-automatic-CL-methods}
\end{center}
\vspace{-6mm}
\end{table*}

\vspace{-2mm}
\subsubsection{Self-Paced Learning}
\label{subsubsec:self-paced-learning}

Self-paced Learning (SPL) is a primary branch of CL that automates the Difficulty Measurer by taking the example-wise training loss of the current model as criteria. The concept of ``self-paced learning'' originates from human education, where the student can control the learning curriculum, including what to study, how to study, when to study, and how long to study~\cite{tullis2011effectiveness}. Under machine learning settings, SPL refers in particular to a training strategy initially proposed by Kumar et al.~\cite{kumar2010self}, 
which trains the model at each iteration with the proportion of data with the lowest training losses.
This proportion of easiest examples gradually grows to the whole training set, which essentially takes a predefined Training Scheduler in Sec.~\ref{subsubsec:predefined-training-scheduler}. 
Note that in the literature of SPL, CL and SPL are usually mentioned as two different strategies, where the CL actually refers to the predefined CL in Sec.~\ref{subsec:4-2-manually-predefined-CL}. 
However, in this paper, SPL is regarded as a branch of automatic CL, since it shares the same spirit with CL and fits perfectly with our general CL framework, as shown in Fig~\ref{fig:CL_methods}(b). 
The most valuable advantages of SPL over predefined CL are mainly two-fold: 1) SPL is semi-automatic CL with a loss-based automatic Difficulty Measurer and dynamic curriculum, which makes it more flexible and adaptive for various tasks and data distributions. 2) SPL embeds the curriculum design into the learning objective of the original machine learning tasks, which makes it widely applicable as a plug-in tool. 



\textbf{a) The Original Version of SPL. }
The original SPL algorithm~\cite{kumar2010self} is formally defined as follows.
Let $\mathcal{D} = \{x_i, y_i\}_{i=1}^N$ denotes the training set, where $x_i$ and $y_i$ is the feature and label of example $i$, respectively. The model $f_{\bm{w}}$ with parameters $\bm{w}$ maps each $x_i$ to the model prediction $f_{\bm{w}}(x_i)$, and gets a loss $l_i = L(f_{\bm{w}}(x_i), y_i)$, where $L$ is the learning objective. The original goal is then to minimize the empirical loss on the whole training set:
\vspace{-2mm}
\begin{equation}
    \footnotesize
	\min \limits_{\bm{w}} \mathbb{E}(\bm{w}; \lambda) \sum_{i=1}^N l_i + R(\bm{w}),
\end{equation}
where $R(\bm{w})$ is a regularizer to encode prior knowledge on $\bm{w}$ to avoid overfitting\footnote{For brevity, we ignore $R(\bm{w})$ in the following discussion.}.
SPL introduces example weight $v_i$ into the above learning objective with an \emph{SP-regularizer} $g(\bm{v};\lambda)$, where $\bm{v} = [v_1, v_2, ..., v_N]^\top \in [0,1]^N$ is a vector of weights, and $\lambda$ is the \emph{age parameter}, a hyperparameter which controls the learning pace (i.e., as Training Scheduler) and determines the proportion of the easiest selected examples at each training epoch. The new learning objective becomes:
\vspace{-2mm}
\begin{equation}
	\label{eqn:SPL-new-objective}
	\footnotesize
	\min \limits_{\bm{w}; \bm{v} \in [0,1]^N} \mathbb{E}(\bm{w}, \bm{v}; \lambda) \sum_{i=1}^N v_i l_i + g(\bm{v};\lambda). \\
\end{equation}
In the original SPL, $g(\bm{v};\lambda)$ is a negative $l_1$-norm: 
\vspace{-2mm}
\begin{equation}
    \footnotesize
	\label{eqn:hard-regularizer}
	g(\bm{v}; \lambda) = - \lambda \sum_{i=1}^N v_i.
\end{equation}
The above learning objective is often optimized with the Alternative Optimization Strategy (AOS)\footnote{AOS is also called ASS (Alternative Search Strategy), ACS (Alternative Convex Search)~\cite{jiang2015self}, or CCM (Cyclic Coordinate Method)~\cite{jiang2014easy} in SPL literature.}. Concretely, we alternatively optimize $\bm{w}$ and $\bm{v}$ while fix the other. With the fixed $\bm{w}^*$, we calculate the global optimum $\bm{v}^*$ by solving:
\vspace{-2mm}
\begin{equation}
	\label{eqn:SPL-update-v}
	\footnotesize
	v_i^* = \arg \min \limits_{v_i \in [0,1]} v_i l_i + g(v_i;\lambda),\quad i=1, 2, \cdots, n\\
\end{equation}
Then, with fixed $\bm{v}^*$, we learn the global optimum $\bm{w}^*$: 
\vspace{-2mm}
\begin{equation}
	\label{eqn:SPL-update-w}
	\footnotesize
	\bm{w}^* = \arg \min \limits_{\bm{w}} \sum_{i=1}^N v_i^* l_i. \\
\end{equation}
The two optimization steps are iteratively conducted, while the value of $\lambda$ is gradually increased to add more harder examples. The overall algorithm is in Algorithm 2. 

\begin{figure}[ht]
\vspace{-5mm}
  \label{alg::original-SPL}
  \renewcommand{\algorithmicrequire}{\textbf{Input:}}
  \renewcommand{\algorithmicensure}{\textbf{Output:}}
  \removelatexerror
  \begin{algorithm}[H]
  \footnotesize
    \caption{Self-Paced Learning}
    \begin{algorithmic}[1]
      \REQUIRE $\mathcal{D} = \{x_i, y_i\}_{i=1}^N$: training dataset; $f$: the machine learning model; $T$: the maximum number of iterations; 
      \ENSURE $\bm{w}$: the optimal parameters of $f$.
      \STATE Initialize $\bm{w}$, $\bm{v}$, $\lambda = \lambda_0$, $t = 0$.
      \WHILE {$t \ne T$}
      \STATE $t = t + 1$;
      \STATE Update $\bm{v}^*$ by Eq.~\ref{eqn:SPL-update-v};
      \STATE Update $\bm{w}^*$ by Eq.~\ref{eqn:SPL-update-w};
      \STATE Update $\lambda$ to a larger value; // to include harder data
      \ENDWHILE
    \end{algorithmic}
  \end{algorithm}
\vspace{-4mm}
\end{figure}

While the solution for Eq.~\ref{eqn:SPL-update-w} 
is provided by machine learning algorithms (e.g., gradient descent) for the original task, the solution for Eq.~\ref{eqn:SPL-update-v} 
is simple. In fact, 
since $g(\bm{v};\lambda)$ in Eq.~\ref{eqn:hard-regularizer} is a convex function of $\bm{v}$, the global minimum can be easily derived by setting the partial derivative of $\mathbb{E}(\bm{w}, \bm{v}; \lambda)$ to $v_i$ as zero. 
Considering $v_i \in [0,1]$, we get the  close-formed optimal solution for $\bm{v}^*$ with the fixed $\bm{w}^*$:
\vspace{-2mm}
\begin{equation}
\label{eqn:binary_v}
    \footnotesize
	v_i^* = \left\{
		\begin{aligned}
			1, & \qquad &l_i < \lambda \\
			0, & \qquad &\text{otherwise}
		\end{aligned}
	\right.
\end{equation}
This solution can be intuitively explained: if an example has a training loss $l_i$ less than the threshold $\lambda$, then it is regarded as an \emph{easy} example for the current model, and should be selected at the current training epoch (i.e., $v_i^* = 1$). Otherwise, it is \emph{hard} and should not be selected (i.e., $v_i^* = 0$). When the model becomes more mature, $\lambda$ gets increased and more harder examples get involved in training. 

Another remaining issue is how to adjust the threshold $\lambda$ 
throughout the training. Initially, $\lambda$ should be set as $\lambda_0$ to ensure that a small proportion of easy examples are selected. Later on, a simple method is to multiply or add a constant at each epoch, i.e., $\lambda_{t+1} = \eta \cdot \lambda_t\ (\eta > 1)$ or $\lambda_{t+1} = \lambda_t + \mu\ (\mu > 0)$, to gradually increase $\lambda$. Finally, $\lambda$ becomes large enough so that all the examples are selected (i.e., $v_i^* = 1\quad \forall i$). This strategy of adjusting $\lambda$ is analogous to predefined continuous Training Scheduler. 
More methods for adjusting $\lambda$ will be discussed in (e).

\textbf{b) Theories for SPL. } 
Before we discuss variant SPL versions enhanced from different aspects, we briefly summarize existing theories on SPL. In short, sound theories have been established for the convergence, robustness, and essence of SPL to support its wide applications.

To begin with, the new learning objective Eq.~\ref{eqn:SPL-new-objective} in SPL is equivalent to the following latent objective function:
\vspace{-2mm}
\begin{equation}
    \footnotesize
	\sum_{i=1}^N F_{\lambda}(l_i) = \sum_{i=1}^N \int_0^{l_i} v_i^*(\tau, \lambda) d\tau
\end{equation}

\noindent where $v_i^*$ is the solution in Eq.~\ref{eqn:SPL-update-v}. Meng et al.~\cite{meng2017theoretical} first prove that the AOS strategy in SPL intrinsically accords with the majorization minimization (MM) algorithm~\cite{lange2000optimization} on a minimization problem of the above latent SPL objective. Therefore, one could leverage theories of MM to provide analyses of the properties of SPL (e.g., convergence). Additionally, they find that this latent objective $\sum_{i=1}^N F_{\lambda}(l_i)$ is also closely related to the non-convex regularized penalty (NCRP), a well-known machine learning methodology with attractive properties in sparse estimation and robust learning, which provides evidence on the robustness of SPL. Based on this work, the authors further prove that the optimization of $\sum_{i=1}^N F_{\lambda}(l_i)$ converges to critical points of the original SPL problem under mild conditions~\cite{ma2018convergence}.

Moreover, Liu et al.~\cite{liu2018understanding} establish a systematic framework for SPL under concave conjugacy theory, which completely tallies with the requirements of SPL models. Based on this framework, they provide a proof for the derived relationship among the SP-regularizer $g(\bm{v};\lambda)$, latent objective $\sum_{i=1}^N F_{\lambda}(l_i)$, and the example weights $\bm{v}$. This result also inspires two general approaches for SPL designs.

\textbf{c) Soft SP-regularizers. }
As a weighting strategy on the learning objective, the core design of SPL is the SP-regularizer $g(\bm{v};\lambda)$, which directly determines the optimal weights $\bm{v}^*$ at each training epoch. Therefore, most of the existing improvements on SPL have been focused on SP-regularizers. 
Recall that in the original version of SPL, $g(\bm{v};\lambda)$ leads to a hard/binary weighting on the examples (Eq.~\ref{eqn:binary_v}), assigning $1$ to easy examples and $0$ to hard examples. However, this style of \emph{hard weights} tends to lose flexibility, since any two ``easy'' (or ``hard'') examples are unlikely to be strictly equally important and learnable~\cite{zhao2015self}. Therefore, an intuitive choice is to design new SP-regularizers to result in \emph{soft weights} $\bm{v}^*$. We call such a group of SP-regularizers \emph{soft regularizers}. 

\begin{table}[htbp]
\vspace{-2mm}
\scriptsize
\caption{Common types of SP-regularizers $g(\bm{v};\lambda)$ and the corresponding close-formed solutions $v^*(l;\lambda)$.}
\vspace{-4mm}
\begin{center}
\begin{tabular}{m{1.6cm}<{\centering}|m{2.9cm}<{\centering}|m{3cm}}
\hline
\textbf{Regularizers} & $g(\bm{v};\lambda)$ & $v_i^*(l_i;\lambda)$ \\
\hline
Hard~\cite{kumar2010self} & $- \lambda \sum_{i=1}^N v_i$ & $\left\{
		\begin{aligned}
			1, & \quad l_i < \lambda \\
			0, & \ \text{otherwise}
		\end{aligned}
	\right.$ \\
\hline
Linear~\cite{jiang2014easy} & $\frac{1}{2}\lambda \sum_{i=1}^N (v_i^2 - 2v_i)$ & $\left\{
		\begin{aligned}
			&1 - l_i / \lambda,  \quad l_i < \lambda \\
			&0,  \ \text{otherwise}
		\end{aligned}
	\right.$\\
\hline
Logarithmic~\cite{jiang2014easy} & 
    \makecell{$\sum_{i=1}^N \left( \zeta v_i - \frac{\zeta^{v_i}}{\log \zeta}\right)$ \\ $\zeta = 1 - \lambda,\ 0<\lambda<1$}
	& $\left\{
		\begin{aligned}
			&\frac{\log (l_i + \zeta)}{\log \zeta},  \quad l_i < \lambda \\
			&0,  \ \text{otherwise}
		\end{aligned}
	\right.$\\
\hline
Mixture~\cite{jiang2014easy} & 
	\makecell{$ -\zeta \sum_{i=1}^N \log\left( v_i + \frac{\zeta}{\lambda_1} \right)$ \\ $\zeta = \frac{\lambda_1 \lambda_2}{\lambda_1 - \lambda_2},$ \\ $\lambda_1 > \lambda_2 > 0$}
	& $\left\{
		\begin{aligned}
			&1, \quad l_i \le \lambda_2 \\
			&0, \quad l_i \ge \lambda_1 \\
			&\zeta\left(\frac{1}{l_i} - \frac{1}{\lambda_1} \right),  \ \text{otherwise}
		\end{aligned}
	\right.$\\
\hline
Mixture2~\cite{zhao2015self} & 
	$\frac{\gamma^2}{v_i + \frac{\gamma}{\lambda}},\quad \gamma > 0$
	& $\left\{
		\begin{aligned}
			&1, \quad l_i \le \left( \frac{\lambda\gamma}{\lambda + \gamma} \right)^2\\
			&0, \quad l_i \ge \lambda^2 \\
			&\gamma\left(\frac{1}{\sqrt{l_i}} - \frac{1}{\lambda} \right),  \ \text{otherwise}
		\end{aligned}
	\right.$\\
\hline
Logistic~\cite{xu2015multi} & 
	\makecell{$\sum_{i=1}^N \text{ln}(\mu_i)^{\mu_i}$ \\ $+ \text{ln}(v_i)^{v_i} - \lambda v_i, \quad \lambda > 0,$ \\
	$\mu_i = 1+e^{-\lambda}-v_i 
	$}
	& $\frac{1+e^{-\lambda}}{1+e^{l_i - \lambda}}$\\
\hline
Polynomial~\cite{gong2018decomposition} & 
	\makecell{$\lambda\left( \frac{1}{t} \sum_{i=1}^N \sum_{i=1}^N v_i \right)$ \\ $\lambda > 0,\ t \in \mathbb{N}^+$}
	& $\left\{
		\begin{aligned}
			&\left(1 - \frac{l_i}{\lambda} \right)^{\frac{1}{t-1}}, \quad l_i < \lambda\\
			&0,  \ \text{otherwise}
		\end{aligned}
	\right.$\\
\hline
\end{tabular}
\label{table:SP-regularizers}
\end{center}
\end{table}

\begin{figure}[htbp]
\vspace{-5mm}
\centerline{\includegraphics[width=0.7\linewidth,height=0.5\linewidth]{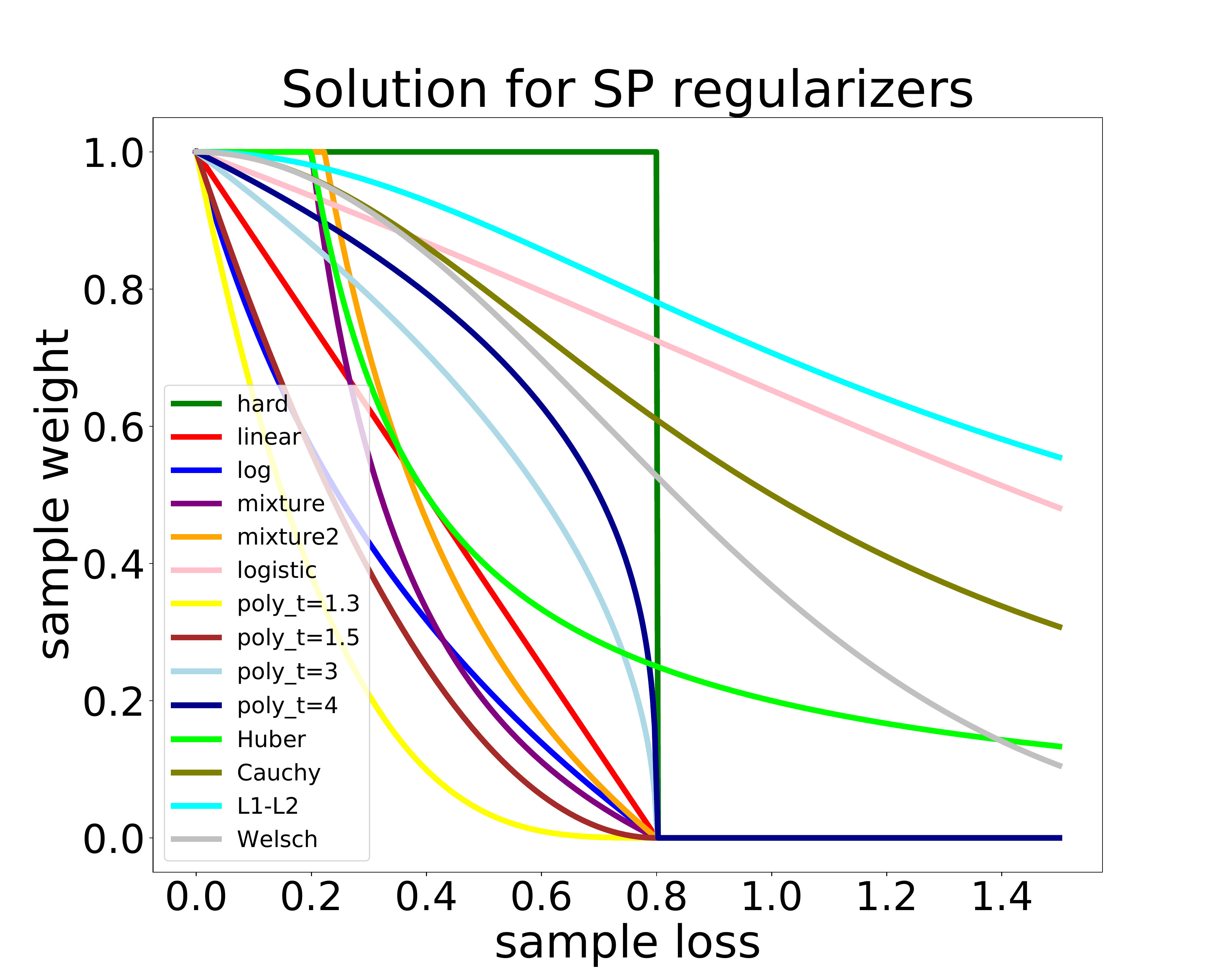}}
\vspace{-2mm}
\caption{Visualization of functions of best example weight $v_i^*$ w.r.t. losses $l_i$ (the $l$-$v^*$ functions) of the SP-regularizers in Table~\ref{table:SP-regularizers}. The age parameter $\lambda$ (the threshold for non-zero weights) for many of the functions are set as $0.8$. The Huber, Cauchy, L1-L2, and Welsch belong to the implicit SP-regularizers in~\cite{fan2016self}, which are not presented in the table.} 
\label{fig:SP_regularizers}
\vspace{-1mm}
\end{figure}

A list of existing SP-regularizers $g(\bm{v};\lambda)$ and the corresponding close-formed solutions of $\bm{v}^*$ is shown in Table~\ref{table:SP-regularizers}. In addition, the $l$-$v^*$ functions (i.e., the function of example weight $v_i^*$ w.r.t. losses $l_i$) of these solutions are visualized in Fig~\ref{fig:SP_regularizers}.
As in Fig~\ref{fig:SP_regularizers}, compared to the hard regularizer, the solutions of various soft regularizers assign soft weights to reflect example importance in finer granularity, which helps soft regularizers achieve better performance in various applications. 
However, one needs to choose suitable soft regularizers for specific scenarios. For example, the \emph{logarithmic} is more prudent than the \emph{linear}, while the \emph{mixture} regularizers tolerate small losses, compared with other regularizers~\cite{jiang2014easy}. The \emph{polynomial} regularizer extends the \emph{linear} to arbitrary orders (when $t=2$, it is identical to \emph{linear}), and Li et al.~\cite{li2017self} further propose to dynamically adjust the order $t$ during training to improve flexibility.

To allow more possibility on SP-regularizer designs, a general and formal definition is taken as follows~\cite{jiang2014easy,zhao2015self}:

\textbf{Definition 4: SP-regularizer}. Suppose that $v$ is a weight variable, $l$ is the loss, and $\lambda$ is the age parameter. $g(\bm{v};\lambda)$ is called a self-paced regularizer, if:

1. $g(v;\lambda)$ is convex w.r.t. $v \in [0, 1]$; 

2. $v^*(l;\lambda)$ is monotonically decreasing w.r.t. $l$, and $\lim\limits_{l\rightarrow0} v^*(l,\lambda)=1, \lim\limits_{l\rightarrow\infty} v^*(l,\lambda)=0$;

3. $v^*(l;\lambda)$ is monotonically increasing w.r.t. $\lambda$, and $\lim\limits_{\lambda \rightarrow \infty} v^*(l,\lambda) \le 1, \lim\limits_{\lambda \rightarrow 0} v^*(l,\lambda)=0$;

\noindent where $v^*(l;\lambda)$ is defined in Eq.~\ref{eqn:SPL-update-v}.

It is not difficult to verify that all the regularizers in Table~\ref{table:SP-regularizers} conform to Definition 4. Based on this definition, Li et al.~\cite{li2017multilabel} propose a general framework for designing SP-regularizers, 
demonstrating that we can derive from any S-shaped $v^*(l;\lambda)$ which meets Conditions 2 and 3 to create new SP-regularizers. 
Essentially, this framework is equivalent to the theorem in~\cite{liu2018understanding}. 

While the SP-regularizers defined by Definition 4 have explicit form, Fan et al.~\cite{fan2016self} further introduce \emph{implicit regularizers} into SPL (denoted as SPL-IR). Based on the convex conjugacy theory, a group of implicit SP-regularizers, whose analytic form can be even unknown, are deduced from some well-studied robust loss functions (e.g., Huber loss function), and the corresponding best weights $\bm{v}^*(l;\lambda)$ can be directly derived from these loss functions. The weights thus inherit the good robustness properties, which helps SPL-IR to outperform explicit SP-regularizers. 
The $l$-$v^*$ functions of implicit regularizers derived from four types of robust loss functions, i.e., Huber, Cauchy, L1-L2, and Welsch loss functions, are visualized in Fig~\ref{fig:SP_regularizers}. \footnote{For clearer comparison with other explicit $l$-$v^*$ functions, we divide the weights $v^*$ by 2 in Cauchy and Welsch. This linear scaling does not influence training if we accordingly amplify the learning rates in SGD.}


\textbf{d) Prior-embedded SPL. }
In SPL methods, given fixed SP-regularizers $g(\bm{v};\lambda)$, the example weights $\bm{v}^*$ are entirely determined by the example-wise losses and the age parameter $\lambda$. However, in some cases, we hope to introduce some \emph{loss prior knowledge} into this learning scheme. For example, we may want to compulsively assign outliers with $v_i = 0$ to improve robustness, or assign pre-known high-quality examples with $v_i = 1$. Such prior knowledge is closely related to the predefined Difficulty Measurer in Sec.~\ref{subsubsec:predefined-difficulty-measurer}. 

Fortunately, the AOS algorithm naturally decomposes SPL into two problems of optimizing $\bm{w}$ and $\bm{v}$, which makes it feasible to embed the loss prior knowledge into SPL by encoding it as a part of SP-regularizer or a constraint on $\bm{v}$. Four typical types of priors are summarized in~\cite{meng2017theoretical} as follows:
i) \emph{Outlier prior}: Some outliers in the datasets show extremely large losses. 
ii) \emph{Spatial/temporal smoothness prior}: Spatially or temporally adjacent examples tend to have similar losses.
iii) \emph{Sample importance order prior}: Some examples are pre-known to be more important than others.
iv) \emph{Diversity prior}: Important examples should be scattered across the data range to help learn global data knowledge.

A famous representative of Prior (iv) is SPL with diversity (SPLD)~\cite{jiang2014self}, which incorporates a negative $l_{2,1}$-norm into the hard SP-regularizer to avoid overfitting to a data subset while ignoring easy examples in other groups: 
\vspace{-2mm}
\begin{equation}
    \footnotesize
	g(\bm{v};\lambda, \gamma) = -\lambda \sum_{i=1}^N v_i - \gamma \sum_{j=1}^b \Vert \bm{v}^{(j)} \Vert_2
\end{equation}
\noindent where $\gamma > 0$ is a balance factor between easiness and diversity, $b$ is the number of groups (e.g., themes in the video event detection task) in the training set, and $\bm{v}^{(j)}$ is a vector of corresponding example weights $v_i$ in group $j$. Since the $l_{2,1}$-norm is well-known to lead to group-wise sparse representation, its opposite term should then encourage diversity of non-zero $v_i$ across groups. 
Alternatively, we can also adopt 
$-l_{0.5,1}$-norm~\cite{zhang2015self}, i.e., $-\sum_{j=1}^b \sqrt{\sum_{i=1}^{n_j}}v_i^{(j)}$, where $n_j$ is the size of group $j$. This diversity term makes the whole $g(\bm{v};\lambda, \gamma)$ conform with Definition 4.
While both $-l_{2,1}$-norm and $-l_{0.5,1}$-norm are based on the Group LASSO~\cite{yuan2006model}, Exclusive LASSO~\cite{kong2014exclusive} can be also adopted~\cite{Han2017SelfpacedMO,ghasedi2019balanced} by taking $-l_{1,2}$-norm to select confident samples from diverse groups or clusters.

For Prior (iii), a representative work is self-paced curriculum learning (SPCL)~\cite{jiang2015self}, which introduces a \emph{curriculum region} $\Psi$ with formal definition as a convex feasible region constraint on $\bm{v}$. SPCL combines the power of SPL and predefined CL, whose objective is as follows:

\vspace{-4mm}
\begin{equation}
    \footnotesize
	\min \limits_{\bm{w}; \bm{v} \in [0,1]^N} \mathbb{E}(\bm{w}, \bm{v}; \lambda, \Psi) \sum_{i=1}^N v_i l_i + g(\bm{v};\lambda).\quad \text{s.t.} \bm{v} \in \Psi \\
\end{equation}
\vspace{-1mm}

An example of $\Psi$ is 
$\{\bm{v} | \bm{a}^{\top} \bm{v} \le c\}$, where $c$ is a constant and $\bm{a}$ is a $N$-dimensional vector derived from the total order relationship among the $N$ examples\footnote{$a_i < a_j$ for all example pairs $(i, j)$ where example $i$ should be learned earlier than example $j$.}. Theoretical analysis on SPCL is provided in~\cite{liu2018understanding}. 

Another method for Prior (iii) is proposed in~\cite{zhang2019leveraging}, which is helpful when the precise total order knowledge is hard to obtain. Similar to the $-l_{2,1}$-norm for Prior (iv), this method encodes the prior knowledge about image difficulty by adding a regularization term $h(\bm{v};\eta, \bm{p}) = -\eta\sum_{i=1}^N p_i v_i$ to the objective, where $p_i$ indicates the priority values of each image. A larger $p_i$ means the example $i$ is easier and should be assigned larger weight $v_i$. 
To generate such $p_i$, all the Difficulty Measurers discussed in Sec.~\ref{subsubsec:predefined-difficulty-measurer} can be adopted. Moreover, SPFTN~\cite{zhang2017spftn} also jointly embeds prior (iii) and (iv) by the weighted sum of terms in~\cite{zhang2019leveraging} and~\cite{zhang2015self}.

Note that when the above kinds of convex constraint on $\bm{v}$ is applied, we could no longer use the close-formed solutions of $\bm{v}^*$ in Table~\ref{table:SP-regularizers}. Instead, we can calculate $\bm{v}^*$ by applying gradient-based methods~\cite{jiang2015self} or other off-the-shelf techniques like CVX toolbox~\cite{zhang2019leveraging} due to the convexity.

\begin{table*}[htbp]
\vspace{-2mm}
\footnotesize
\caption{Representatives of Transfer Teacher. Diff. = different.}
\vspace{-4mm}
\begin{center}
\begin{tabular}{p{3.4cm}p{4.2cm}p{4.6cm}p{3.8cm}}
\hline
\textbf{Representatives} & \textbf{Teacher model} & \textbf{Teacher pretraining dataset} & \textbf{Difficulty} \\
\hline
Transfer learning~\cite{weinshall2018curriculum} & Diff. structure with student & ImageNet & Loss \\
Bootstrapping~\cite{hacohen2019power} & Same structure as student & The training dataset & Loss \\
Cross Review~\cite{xu2020curriculum} & Same structure as student & $N$ training subset & Loss \\
Uncertainty~\cite{zhang2018empirical,zhou2020uncertainty} & Language model & The training dataset & Cross entropy \\
Domain score~\cite{zhang2019curriculum,wang2019dynamically} & Language model & General- and in-domain datasets & Cross entropy difference \\
Noise score~\cite{wang2019dynamically} & Same NMT models as student & Noisy and clean datasets & Cross entropy difference \\
\hline
\end{tabular}
\label{table:Transfer-Teachers}
\end{center}
\vspace{-5mm}
\end{table*}

\textbf{e) Other enhancements of SPL. }
Besides the various enhanced versions of SP-regularizers, there remain some other aspects to be carefully considered in SPL. 
A key element in SPL is the age parameter $\lambda$. 
As aforementioned, traditional SPL takes a naive strategy to add/multiply $\lambda$ with a constant at each epoch. However, with the model making progress, the losses on all the examples are expected to become smaller and smaller, and thus an monotonic increasing threshold $\lambda$ may add much more hard examples in the early epochs. For some SP-regularizers it would be more effective to gradually decrease the value of $\lambda$~\cite{fan2016self}. To design a better update strategy for $\lambda$, some works~\cite{ren2017robust,li2017self} adopt a strategy analogous to Baby Step scheduler in Sec.~\ref{subsubsec:predefined-training-scheduler}. They predefine a sequence $\bm{N} = \{N_1, N_2, \cdots, N_T\}$ ($N_s < N_t$ for all $s < t$, $N_T = N$), where $N_t$ is the number of selected examples in the $t$-th epoch. Then, the threshold of $\lambda$ is dynamically updated to ensure exactly $N_t$ examples are assigned with non-zero weights $v_i$ in the $t$-th epoch. Lin et al.~\cite{lin2017active} also propose to adjust $\lambda$ as follows: 
\vspace{-2mm}
\begin{equation}
    \footnotesize
	\lambda_t = \left\{
		\begin{aligned}
			& \lambda_0, \quad t = 0&\\
			& \lambda_{t-1} + \alpha \cdot \eta_{t}, \quad 1 \le t \le \tau&\\
			& \lambda_{t-1}, \quad t > \tau,&
		\end{aligned}
	\right.
\end{equation}
\noindent where $\eta_{t}$ is the model performance (e.g., accuracy) in the $t$-th epoch. When $\eta_{t}$ is high, then $\lambda$ will increase by a bigger step to add more harder examples, and vice versa. Recently, Shu et al.~\cite{shu2020meta} further propose to leverage meta-learning paradigm to optimize $\lambda$ based on a small and high-quality valid set, which entirely automates the update of $\lambda$. 

In addition to $\lambda$, other hyperparameters, including initialization and stopping criteria, are also very difficult to determine and heavily influencing the SPL performance. What is more, each configuration of hyperparameters could only lead to a single solution, losing view for the entire solution spectrum~\cite{li2016multi}. To address these issues, Li et al.~\cite{li2016multi,gong2018decomposition} propose to discard the traditional AOS algorithm and reformulate the SPL problem as a multi-objective issue, 
which can obtain a set of solutions with different stopping criteria in a single run and improve the robustness of SPL even under bad initialization.


\textbf{f) Applications of SPL. }
SPL has been widely applied to many practical problems, including CV tasks of visual category discovery~\cite{lee2011learning}, segmentation learning~\cite{kumar2011learning,zhang2017spftn}, image classification~\cite{tang2012self}, object detection~\cite{tang2012shifting,zhang2019leveraging}, reranking in multimedia retrieval~\cite{jiang2014easy}, person ReID~\cite{zhou2018deep} etc., and traditional machine learning tasks of matrix factorization~\cite{zhao2015self}, feature selection~\cite{zheng2020unsupervised}, cross-modal matching~\cite{liang2016self}, co-training~\cite{ma2017self}, clustering~\cite{xu2015multi,ghasedi2019balanced,yu2020self}, etc. 
As a primary branch of CL, SPL has the same application motivations as CL, i.e., to guide and to denoise (see Sec.~\ref{subsec:3-2-application-scenes}). Besides, SPL is also effective for a group of applications where the algorithm needs to assign pseudo-labels by models, including reranking~\cite{jiang2014easy}, co-saliency detection~\cite{zhang2015self}, and other weakly~\cite{Han2019WeaklySupervisedLO} or unsupervised learning tasks~\cite{ghasedi2019balanced}. 
Additionally, 
some works also extend SPL by introducing group-wise weights to improve the performance on multiple data groups, e.g., multi-modal~\cite{gong2016multi}, multi-view~\cite{xu2015multi}, multi-instance~\cite{zhang2015self}, multi-label~\cite{li2017multilabel}, multi-class~\cite{ren2017robust}, multi-task~\cite{li2016multitask}, etc. Finally, SPL is also combined with complementary data-selection-based training strategies like boosting~\cite{pi2016self} and active learning~\cite{lin2017active,tang2019self} to benefit both schemes.
Beyond the scope of SPL, the idea of ``deciding learning materials by student'' has also inspired self-paced-like designs in broader contexts, e.g., contextual RL~\cite{Klink2019SelfPacedCR}, knowledge distillation~\cite{Xiang2020LearningFM}, etc.


\vspace{-2mm}
\subsubsection{Transfer Teacher}
\label{subsubsec:transfer-teacher}

SPL takes the current student model as an automatic Difficulty Measurer. 
However, this strategy has a risk of uncertainty at the beginning of training, when the student model is not mature enough (i.e., not sufficiently trained). 
This is analogous to human education: if a student understands little about the learning materials, it would be hard for him/her to measure the difficulty of the materials and find out the easy ones. Thus, a natural idea is to invite a mature teacher to help the student assess the materials and form an easy-to-hard curriculum. This idea leads to the CL approaches that we denote as Transfer Teacher.
As illustrated in Fig~\ref{fig:CL_methods}(c), it is a semi-automatic CL method. 
Particularly, this method first pretrains a teacher model on the training dataset or an external dataset (e.g., ImageNet), and then 
transfers its knowledge to calculate the example-wise difficulty, based on which a predefined Training Scheduler can be applied to finish the CL design. 
Transfer Teacher reduces the burden of artificial Difficulty Measurer designs and thus could be helpful to the tasks where the example-wise easiness is hard to measure. 


Some representatives of Transfer Teacher are presented in Table~\ref{table:Transfer-Teachers}. The most general Transfer Teachers are the loss-based methods (the first three rows), which do not need any domain knowledge and are closely related to SPL. Concretely, these methods take the example-wise losses calculated by a teacher model as the example difficulty and assume that the lower the loss, the easier the example. The teacher model can either be different from the student model and have the greater model capacity (i.e., more complex)~\cite{weinshall2018curriculum}, or share the same structure with the student model~\cite{hacohen2019power,xu2020curriculum}. For instance, in~\cite{weinshall2018curriculum}, a strong teacher classifier pretrained on ImageNet is taken to transfer its knowledge to calculate the example-wise losses on the training dataset. 
The authors in~\cite{hacohen2019power} adopts a bootstrapping 
strategy, which uses a teacher classifier with the same network structure as the student classifier, and pretrains it on the training dataset. This pretrained teacher can be regarded as a mature version of the student to calculate loss-based difficulty. Note that the difference between bootstrapping and SPL is that the former's Difficulty Measurer is mature and fixed, while the latter's is the current student model which gradually grows up. 
Another example of loss-based Transfer Teacher is the Cross Review strategy~\cite{xu2020curriculum}, which alleviates the fluctuation of the difficulty measurement. 
Concretely, the authors uniformly divide the trainset into $N$ shares and train one teacher on each share. Then for each example in the $i$-th share, they take the other $N-1$ teachers to calculate a loss-based difficulty score. 

\begin{table*}[htbp]
\caption{Representatives of RL Teacher. Acc. = accuracy, thres. = threshold.}
\vspace{-4mm}
\begin{center}
\begin{tabular}{p{3cm}p{3.7cm}p{5.8cm}p{2.5cm}}
\hline
\textbf{Representatives} & \textbf{RL Algorithm} & \textbf{Reward/Student Feedback} & \textbf{Main Goal} \\
\hline
AutoCL~\cite{graves2017automated} & Multi-armed bandit & Loss/Complexity-driven learning progress & Efficiency \\
TSCL~\cite{matiisen2019teacher} & Non-stationary bandit & Absolute value of slope of learning curve & Efficiency \\
L2T~\cite{fan2018learning} & REINFORCE & How fast the student achieve valid acc. thres. & Efficiency \\
RL-based CL~\cite{kumar2019reinforcement} & Q-Learning & Log-likelihood on valid set & Performance \\
RCL~\cite{zhao2020reinforced} & Discriministic Actor-Critic & Perplexity difference on valid set & Performance \\
\hline
\end{tabular}
\label{table:RL-Teachers}
\end{center}
\vspace{-5mm}
\end{table*}

Moreover, in NLP literature, there exist some typical methods adopting a ``teacher'' model to measure example-wise difficulty for training data selection, which can be naturally incorporated into CL as Transfer Teacher. 
For example, some works~\cite{zhang2018empirical,zhou2020uncertainty} leverage the following model-based data uncertainty $u^{data}(s) = - \frac{1}{|s|} \sum_{i=1}^{|s|} \log P(s_i | s_{<i})$ to measure sentence-wise difficulty in NMT tasks, 
where $P(s_i | s_{<i})$ is the confidence of the pretrained language model (LM) for its prediction about the $i$-th word in sentence $s$, and $|s|$ is the length of $s$. The lower of this uncertainty score, the easier the sentence according to the teacher LM.
Besides, Moore et al.~\cite{moore2010intelligent} propose to use two LMs to measure how much a sentence $s$ is related to a specific domain (e.g., news, talks, patents, etc.) and select domain sentences. 
This measurement of domain score is leveraged in~\cite{zhang2019curriculum,wang2019dynamically} as Transfer Teacher 
according to the specific scenarios (e.g., in-domain data can be seen as easier for domain adaption).
Moreover, Wang et al.~\cite{wang2019dynamically} also use two NMT models to measure the noise level of a sentence pair $\{x, y\}$. 
A lower noise level refers to cleaner and also easier data.

\vspace{-1mm}
\subsubsection{RL Teacher}
\label{subsubsec:RL-teacher}

The SPL and Transfer Teacher only automate the Difficulty Measurer and still use predefined Training Scheduler, and they only consider one side of the ``curriculum'' or teaching scenario: SPL takes the student feedback (i.e., losses) to adjust the curriculum, while Transfer Teacher leverages the teacher's knowledge to determine the order of presenting learning materials. A common sense in human education is that an ideal teaching strategy should involve both the teacher and the student, where the student could interactively provide feedback to the teacher, and the teacher could then adjust the teaching action accordingly. In this way, both the teacher and student will make progress together. 

To this end, RL Teacher methods are proposed, which involve a student model and a reinforcement-learning-based teacher model. At each training epoch, the RL teacher will dynamically select examples/tasks for training according to the student feedback. Concretely, the data selection is taken as the \emph{action} in the RL schemes, and the student feedback is taken as the \emph{state} and \emph{reward}. From the view of the general CL framework in Sec.~\ref{subsec:4-1-general-framework}, 
the RL Teacher sets the teacher model as both the Difficulty Measurer and Training Scheduler by dynamically considering the student feedback. 
The illustration of RL Teacher is shown in Fig~\ref{fig:CL_methods}(d). It is clear to see that, with this teacher-student interactive strategy, RL Teacher achieves the fully-automated CL design. 


Some representatives of the RL Teacher are listed in Table~\ref{table:RL-Teachers}. Both traditional RL and deep RL models are leveraged in these designs, where the deep RL models are stronger in performance but more time-consuming and harder to train. It is worth mentioning that RL Teacher methods make it possible to set different student feedback according to different goals, e.g., training efficiency or generalization performance, which brings great flexibility and applicability to various scenarios. Additionally, RL Teacher is typically suitable for multi-task learning, where the teacher model selects the most valuable tasks for the student training.

AutoCL~\cite{graves2017automated} and TSCL~\cite{matiisen2019teacher} are two RL Teacher methods designed for multi-task settings, where the goal is to learn a student model that achieves high performance on all the tasks. In both works, bandit-based RL models are adopted as the teacher model, whose job is to receive the reward signals from the student model and select one training task for student learning in the next epoch. Specifically, the RL teachers learn the mapping from history reward sequences $\bm{r} = \{r_i\}_{i=1}^N$ (of different tasks) to the probability vector $\bm{\pi}$ of sampling the $N$ training tasks. 
As both the works aim to design a CL algorithm to improve the training efficiency, various reward measurements are proposed. In AutoCL~\cite{graves2017automated}, the authors define a group of \emph{learning progress} as the reward, which includes loss-driven and complexity-driven measurements. The intuition is, if a decrease in some loss or an increase in the student model's complexity is observed after training on the $i$-th task, then this task is helpful to the student model for making big progress and should be assigned larger sampling probability. 
On the other hand, in TSCL~\cite{matiisen2019teacher}, the authors set the reward as the absolute value of the slope of the learning curve (the absolute difference between the performance scores of two successive epochs) 
on a specific task. This is an elegant design: when the slope is a large positive value, it means the student is making progress on this task; and when the slope is a large negative value, it implies that the student is forgetting this task. Both conditions should lead to a larger sampling probability on this task to achieve faster and more generalizable student training.

L2T (Learning to Teach)~\cite{fan2018learning} adopts the REINFORCE algorithm as the RL teacher. Given a random mini-batch $D_t$ in the $t$-th supervised training epoch, the goal of the teacher model is to dynamically determine which data examples are used and which are abandoned. To this end, the action $\bm{a}_t = \{a_t^{(m)}\}_{m=1}^M \in \{0,1\}^M$ is a hard selection on each of the $M$ examples in this mini-batch. The state $s_t = (D_t, f_t)$ is defined as the concatenation of various features of the current mini-batch $D_t$ and the current state of student model $f_t$\footnote{For example, data features include the predefined Difficulty Measurer features in Table~\ref{table:difficulty-measurer}, and model features include iteration number, average historical training loss / validation accuracy, etc.}. This design of state/observation is quite general and applicable to most learning scenarios. 
Moreover, aiming at fast convergence, the reward $r_t$ is set as a terminal reward (i.e., $r_t = 0, \forall t < T$) to be related with how fast the student model learns. In particular, $r_T = -\log(i_{\tau} / T')$, where $i_{\tau}$ is the iteration number for the student model achieving an accuracy threshold $\tau \in [0, 1]$ on the valid set, and $T'$ is a predefined maximum iteration number. With all the definition above, L2T trains the teacher model by maximizing the expected reward $J(\theta) = \mathbb{E}_{\phi_{\theta}(a|s)}[R(s,a)]$, 
where $R(s,a)$ is a state-action value function to estimate the reward, and $\phi_{\theta}$ is the data selection policy parameterized by $\theta$, which can be any binary classification model. Through this dynamic data selection by the teacher model, the student model is expected to converge faster to a better optima.

\begin{table*}[htbp]
\caption{Representatives of ``Other Automatic CL'' automatic CL methods.}
\vspace{-4mm}
\begin{center}
\begin{tabular}{p{5cm}p{5cm}p{4cm}}
\hline
\textbf{Papers} & \textbf{What to Optimize} & \textbf{How to Optimize} \\
\hline
Learning CL with BO~\cite{tsvetkov2016learning} & Weights for difficulty dimensions & Bayesian Optimization \\
MentorNet~\cite{jiang2018mentornet}, ScreenerNet~\cite{kim2018screenernet} & Loss weights & SGD  \\
APL~\cite{Zhang2020FewCostSO} & Loss weights & Adversarial learning \\
Learning to reweight~\cite{ren2018learning} & Loss weights & Meta-learning\\
L2T with dynamic loss function~\cite{wu2018learning} & Loss function (as a linear model) & Hypernetwork \\
Data Parameters~\cite{saxena2019data} & Class/Instance-wise loss function & Data Parameters \\
\hline
\end{tabular}
\label{table:other-automatic-CL}
\end{center}
\vspace{-5mm}
\end{table*}

Beyond traditional RL algorithms, recent works also leverage deep RL models, e.g., Q-learning~\cite{kumar2019reinforcement} and Deterministic Actor-Critic~\cite{zhao2020reinforced}, to design RL Teacher methods for automatic data selection, sharing the same spirit with L2T.  Both the two works focus on the NMT task, a typical application for CL discussed in Sec.~\ref{subsec:3-2-application-scenes}. RL-based CL~\cite{kumar2019reinforcement} first sorts the examples according to a predefined measurement and divide them into $M$ bins of equal sizes, and then defines the action as selecting one bin for NMT training. The reward and state are related to the log-likelihood on the valid set and a prototype batch sampled from all bins, respectively. Moreover, in RCL~\cite{zhao2020reinforced}, the state $s$ is similarly defined as L2T, including feature embeddings from data and the student model. Given $s$, the actor network $\mu$ is optimized to select examples from a mini-batch (i.e. action $a = \mu(s)$) to form the training set at each epoch, such that the estimated reward $Q(s, a)$ by critic network $Q$ is maximized. The critic network, on the other hand, is optimized to estimate the reward $r$ more accurately, where $r$ is defined as the performance improvement of the student model on the valid set after trained. Compared with traditional RL methods like REINFORCE, Actor-Critic is supposed to help reduce the update variance and accelerate convergence.

\subsubsection{Other Automatic CL}
\label{subsubsec:other-automatic-CL}

Besides RL Teacher, there exist some other fully-automatic CL designs. Intuitively, these designs should require the generation of the curriculum to rely only on the dataset, the student model, and the goal of the task. According to the CL definition in Sec.~\ref{sec:2-definition}, we can regard this curriculum as a sequence of training criteria or objectives. Thus, from the optimization perspective, 
at each training epoch, we hope to optimize the following mapping to improve performance: $ \{\text{data, current state of student model, task goal}\} \mapsto \text{training objective} $.
To this end, RL Teacher methods typically adopt an RL framework to learn the policy for training data selection. Additionally, more optimization methods, such as Bayesian Optimization (BO), Stochastic Gradient Descent (SGD), Meta-learning, and Hypernetwork, are also demonstrated to have great potential to learn this mapping. 
Note that these methods can also be regarded as a ``teacher'' searching for the best curriculum according to the student state/feedback. 
Since the methodologies and focuses of optimization are diverse in these works, we conclude them in this subsubsection as ``Other Automatic CL'' (Table~\ref{table:other-automatic-CL}).

Tsvetkov et al.~\cite{tsvetkov2016learning} make one of the earliest attempts on automatic CL by leveraging BO to learn the best curricula for word representation learning. 
The curriculum here is determined by the scalar product of a learned weight vector $\bm{w}$ and an example-wise difficulty feature vector $\bm{x}$, according to which the examples are scored and sorted for later representation learning. While $\bm{x}$ is manually engineered, 
the weight vector $\bm{w}$ learned by BO provides the possibility for different curriculum according to different downstream tasks. Specifically, BO in this work is a sequential approach to performing a regression from $\bm{w}$ to the performance on the downstream task. At the $t$-th iteration, the algorithm first sort the examples by the $\bm{w}_{t} \cdot \bm{x}$, learn word representations $V_t$ (i.e., student model) with this curriculum, and then train extrinsic models on downstream task and evaluate the performance $eval_t$. Finally, $eval_t$ is collected by BO algorithm to generate the $\bm{w}_{t+1}$. Through this process, BO learns to predict a better $\bm{w}$ and thus a better curriculum.

While SPL methods in Sec.~\ref{subsubsec:self-paced-learning} optimize the example-wise loss weights $\bm{v}$ by solving the new objective with manually designed SP-regularizers, existing works have made further effort to optimize $\bm{v}$ throughout training by different approaches. One idea is to predict the loss weight $v_i$ of example $\{x_i, y_i\}$ by a teacher model, which is adopted in MentorNet~\cite{jiang2018mentornet} and ScreenerNet~\cite{kim2018screenernet}. The MentorNet $h$ is a teacher model with parameters $\Theta$ which maps the example-wise feature $z_i = \phi(x_i, y_i, \bm{w})$ to the corresponding loss weight $v_i$. Here, $z_i$ includes the loss, loss difference to the moving average, label, and epoch percentage, and $\bm{w}$ denotes the parameters of the student model. Given fixed $\bm{w}$, the MentorNet is trained on a trusted small dataset $\mathcal{D}_{val}$ by SGD:
\vspace{-2mm}
\begin{equation}
    \footnotesize
	\Theta^* = \arg \min \limits_{\Theta} \sum_{i \in \mathcal{D}_{val}} \text{CE} (h(z_i; \Theta), v_i^*),
\vspace{-1mm}
\end{equation}
\noindent where $v_i^*$ is manually annotated as $1$ iff $y_i$ is a correct label and $0$ otherwise, and CE stands for cross-entropy. During the mini-batch training of the student model, the MentorNet is only updated a fixed number of times (with student fixed). Besides the data-driven curriculum learned on $\mathcal{D}_{val}$, we could also train the MentorNet to approximate a predefined curriculum, e.g., by setting $v_i^*$ as the loss weights derived from some SPL objectives. The convergence and robustness of student learning are also theoretically proved.




Apart from teacher model, APL~\cite{Zhang2020FewCostSO} also predicts the loss weights $\bm{v}$ in SPL by generative adversarial learning. Concretely, under semi-supervised setting, a pace-generator $P$ outputting $\bm{v}$ is trained to discriminate annotated ($v_i=1$) and predicted ($v_i=0$) labels, and a task-predictor $T$ predicting labels is alternatively trained with $P$ to produce high-quality predictions. After the initial training on labeled data, the unlabeled data is then added to the training set with loss weights (or ``pace'') $\bm{v}$ given by $P$ in each iteration. This APL paradigm is proven significantly more effective than SPL methods on the task of salient object detection with few labeled data. An analogous idea is adopted in~\cite{Hung2018AdversarialLF} by assigning binary selection on unlabeled data based on pretrained discriminator on labeled data in semi-supervised semantic segmentation task.

Ren et al.~\cite{ren2018learning} further propose a meta-learning~\cite{hospedales2020meta} perspective for optimizing loss weights $\bm{v}$. 
Akin to MentorNet, a clean unbiased valid set is adopted to guide the meta-learning. Specifically, at the $t$-th epoch, they first locally update the student model (with parameters $\bm{w}_t$) by one gradient step on a training mini-batch $\mathcal{D}_{train}$, where the example weights $v_i$ are perturbed by $\epsilon_i$:
\vspace{-2mm}
\begin{equation}
    \scriptsize
	\hat{\bm{w}}_{t+1}(\epsilon) = \bm{w}_t - \alpha \nabla \sum_{i\in \mathcal{D}_{train}} \epsilon_i l_i(\bm{w}_t),
\end{equation}
\noindent where $l_i(\bm{w})$ is the loss and $\alpha$ is the local learning rate. To estimate the best loss weights $\bm{v}$ according to the clean valid set, they take a meta-gradient step on a validation mini-batch $\mathcal{D}_{val}$ w.r.t. $\epsilon$, and force the weights to be non-negative:
\vspace{-2mm}
\begin{equation}
    \scriptsize
	\tilde{v}_{i,t} = \max \left( 0, -\eta \frac{\partial }{\partial \epsilon_j} \frac{1}{|\mathcal{D}_{val}|} \sum_{j \in \mathcal{D}_{val}} \hat{\bm{w}}_{t+1}(\epsilon) \right) ,
\end{equation}
\noindent where $\eta$ is the meta learning rate. The $\tilde{\bm{v}}_t$ is then normalized to obtain the final new weights $\bm{v}_t$. Finally, they meta update the model parameters to $\bm{w}_{t+1}$ with the new objective weighted by $\bm{v}_t$, i.e., $\sum_{i\in \mathcal{D}_{train}} v_{i,t} l_i(\bm{w}_t)$. 
This meta-learning mechanism would lead the student model to converge to an appropriate distribution favored by the clean and balanced valid set and thus become more generalizable and robust.

Beyond loss weights, some other works~\cite{wu2018learning,saxena2019data} also focus on learning dynamic loss function as a whole, which complies with the most general definition of CL in Sec.~\ref{sec:2-definition}. As argued in L2T~\cite{fan2018learning}, while data selection is analogous to human teacher selecting teaching materials, designing good loss function corresponds to human teacher determining the examination criteria, which is another significant issue in a ``curriculum''. In~\cite{wu2018learning}, the scholars propose to leverage a two-layer perceptron as the teacher hypernetwork $\mu_{\Theta}$ to predict the parameters of the loss function $l_{\Phi}(\hat{y},y)$. In other words, the loss function is assumed to be itself a neural network with coefficients $\Phi$, and at the $t$-th epoch, $\Phi_t = \mu_{\Theta}(s_t)$, where $s_t$ is the state vector of the student model $f_{\bm{w}}$. Akin to MentorNet, the goal of the teacher model is to maximize the performance of induced student model on a valid set $\mathcal{D}_{val}$: $\Theta^* = \max \limits_{\Theta} \mathcal{M} (f_{\bm{w}^*}, \mathcal{D}_{val})$, 
where $f_{\bm{w}^*} = \mathcal{F}(\mathcal{D}_{val}, \mu_{\Theta})$ stands for the student model trained on the training set with the loss function predicted by $\mu_{\Theta}$, and $\mathcal{M}$ is the performance measurement on $\mathcal{D}_{val}$. Novel algorithms are also proposed to make this optimization of teacher hypernetwork possible.




\subsection{How to Choose A Proper CL Method}
\label{subsec:4-4-choose-my-CL}

Although we have reviewed the major ideas of different CL methodologies, how to choose them in real-world applications remains an important problem, which is rarely discussed in existing CL literature and there is no systematic conclusion. In this subsection, we make effort to summarize some empirical evidence and ideas on this topic.

\textbf{Conclusions from empirical studies.} Although such work is scarce, different CL methods are still compared and analyzed in a small number of works. Cirik et al.~\cite{cirik2016visualizing} compare different predefined schedulers on two sequence prediction tasks with LSTM models, showing that predefined CL benefits more when smaller models are applied and the size of the training set is limited. Zhang et al.~\cite{zhang2018empirical} experiment on the combinations of various predefined Difficulty Measurer and various predefined Training Scheduler on neural machine translation task, reaching the result that predefined CL is highly sensitive to the choices of Difficulty Measurer and hyperparameters (i.e., learning rates). Hacohen et al.~\cite{hacohen2019power} compare SPL, anti-curriculum, and different Transfer Teacher methods with various Training Schedulers on image classification, demonstrating Transfer Teacher is the most robust, and the advantage of CL is more effective when the task is difficult. 

Within each CL category, several empirical conclusions can be summarized as follows ($>$ means more effective), although most of them are not universal. 
(i) Predefined CL: for Training Scheduler, the continuous root-$p$ function $> $ discrete Baby Step $>$ discrete One-Pass (see Fig.~\ref{fig:continuous_schedulers})~\cite{penha2019curriculum,platanios2019competence,cirik2016visualizing}.
(ii) SPL: for SP regularizers without embedded prior, implicit regularizers $>$ soft regularizers (e.g., \emph{mixture, logarithmic}) $>$ hard regularizers (see Fig.~\ref{fig:SP_regularizers})~\cite{fan2016self,jiang2014easy}.
(iii) SPL: if reliable prior knowledge or assumption is available, embedding it into the SPL objective always help~\cite{jiang2014self,jiang2015self,ghasedi2019balanced,zhang2019leveraging}.
(iv) Fully Automatic CL (Sec.~\ref{subsubsec:RL-teacher},~\ref{subsubsec:other-automatic-CL}): Many fully automatic CL methods are shown to be significantly more effective than SPL methods on weakly-supervised CV and NLP tasks~\cite{jiang2018mentornet,ren2018learning,saxena2019data,Zhang2020FewCostSO,wang2020optimizing}, while MentorNet~\cite{jiang2018mentornet} is often selected as a baseline in these papers.

The best selection among different CL categories needs further empirical studies. However, qualitative comparison of different methodologies is provided in Table~\ref{table:comparison-of-predefined-and-automatic-CL} and~\ref{table:comparison-of-automatic-CL-methods}. A principle for selecting a proper CL category is to consider how much prior knowledge you know about your dataset and task goal. If sufficient expert domain knowledge is available, then predefined CL methods are more preferable to design a \emph{knowledge-driven} curriculum specifically suitable to the exact scenario. On the other hand, if we have no prior assumptions on the data, then automatic CL methods are more preferable to learn a \emph{data-driven} curriculum adaptive to the underlying dataset and task goal.

\textbf{Hybrid CL.} A further consideration of designing a CL framework is to adopt different CL methods jointly, making them complement each other.  Generally, this hybrid CL can be designed by applying different CL methods on different evidence for curriculum or different levels of data. 
A typical example is the SPCL-like methods~\cite{jiang2015self,zhang2019leveraging,zhang2017spftn,Zhang2019LearningOD} in Sec.~\ref{subsubsec:self-paced-learning} that embed the predefined sample-importance-order prior into the SPL objectives or SPL-like regimes, taking the advantages of both knowledge-driven predefined CL and data-driven SPL to enrich the curriculum from both sources of evidence, i.e., human and machine. Following this paradigm, an interesting idea for future researchers might be to embed human prior on sample importance into the fully data-driven CL methods in Sec.~\ref{subsubsec:RL-teacher} and~\ref{subsubsec:other-automatic-CL}, which is being explored by frontier researchers~\cite{wang2020optimizing}. 
On the other hand, we can also apply different CL to different levels of training data. For example, LFME~\cite{Xiang2020LearningFM} jointly adopts an SPL-like mechanism for expert selection (each expert is trained on a subset of training data) in knowledge distillation and a Transfer Teacher for instance selection in each subset.


\textbf{Extra computational cost of CL.}
It is worth mentioning another concern of great practical significance: though seemingly effective and easy-to-use, how much does it cost to apply these CL methods, i.e., the extra computational cost to the training? 
Before the analysis on the time complexity of CL, we remind readers that convergence speedup is one of the main advantages and motivations of CL, and many CL methods in different categories (e.g.,~\cite{platanios2019competence,jiang2014self,fan2018learning}) can actually accelerate training. By reducing the number of iterations to convergence, the total cost of training is reduced despite additional computations for CL.

As additional computational complexity of CL is hardly discussed in the literature, we generally analyze it according to the taxonomy in Sec.~\ref{sec:4-CL-design-a-general-framework}. We assume there are $n$ training examples to train $M$ iterations. 
(i) Predefined CL methods in Sec.~\ref{subsec:4-2-manually-predefined-CL} calculate and then fix the curriculum before the training process starts. It often costs $O(n)$ (or $O(1)$ if human annotation is available) to calculate the difficulty of each sample and $O(n\log n)$ to sort the samples from easy to hard. During training, the scheduler calculates the difficulty threshold for batch sampling at each iteration, which costs $O(1)$ (for discrete schedulers) or $O(M)$ (for continuous schedulers, see Sec.~\ref{subsubsec:predefined-training-scheduler}). Thus, the overall complexity is $O(n\log n + M)$, which is the cheapest among all CL methods.
(ii) SPL methods in Sec.~\ref{subsubsec:self-paced-learning} dynamically updates the sample weights $\bm{v}=\{v_i\}_1^n$ at each iteration, and thus the extra complexity is $O(Mn)$ or $O(Mnx)$ if close-formed solutions of $\bm{v}^*$ exists or not, where $x$ is the computations of CVX toolbox for convex optimization on $v_i$.
(iii) Transfer Teacher methods in Sec.~\ref{subsubsec:transfer-teacher} pretrain a teacher difficulty measurer before training, then it calculates a curriculum like predefined CL. So the overall complexity is $O(T + n\log n + M)$, where $T$ is the cost of pretraining the teacher. 
(iv) RL Teacher methods in Sec.~\ref{subsubsec:RL-teacher} dynamically learn the data weighting policy of the teacher and learn the student model at each iteration. The overall complexity is $O(RM + xMn)$, if $R$ is the computations for one updating step of teacher, and $x$ for predicting the weight for one example. $R$ can be both small (bandit) and large (Deep RL).

In summary, from the theoretical perspective of time complexity, most CL methods in (i) to (iv) induce little or acceptable additional cost w.r.t. the cost of main training and are thus worth adoption according to their advantages. We have to admit that CL can also be expensive, e.g., Deep RL Teacher~\cite{kumar2019reinforcement,zhao2020reinforced}. Generating a task curriculum in RL setting often costs greater time than learning the tasks~\cite{narvekar2020curriculum}. However, it is a trade-off between performance and efficiency, e.g., RL agents may fail to solve the target tasks without the expensive curriculum~\cite{florensa2017reverse}.

\vspace{-3mm}
\section{Discussions}
\label{sec:5-discussions}



\vspace{-2mm}
\subsection{Easier First v.s. Harder First}
\label{subsec:5-1-easier-first-vs-harder-first}

A fundamental question for the CL strategy (in Definition 1) is: does this ``easy to hard'' training strategy always help, given all of these works and theories? In some literature of CL, the answer to this question is ``No''. For example, Avramova~\cite{avramova2015curriculum} finds that convolutional neural networks derive most learning values from the hardest examples, and the damage of excluding those easiest examples is minor. 
Zhang et al.~\cite{zhang2018empirical} also test a reverse version of CL (i.e., a copy of baseline CL reversing the difficulty ranking to ``hard to easy'', also called \emph{anti-curriculum}), on NMT tasks, which shows that in some cases, anti-curriculum may even achieve the best performance among various Training Scheduler designs. Besides, Hacohen et al.~\cite{hacohen2019power} demonstrate that SPL will hurt the performance and significantly delay learning in their experiments. Other works~\cite{zhou2018minimax,wang2019dynamically} also design ``harder examples first'' curricula.

Besides CL literature, hard example mining (HEM)~\cite{shrivastava2016training} serves as another well-studied and popular data selection strategy, which is opposite to CL. Concretely, in each training batch, HEM selects the hardest examples for training (or assign them with higher weights), assuming that the harder examples are more informative. The difficulty in HEM is often defined according to the current model losses on examples~\cite{shrivastava2016training,loshchilov2015online} or the gradient magnitude~\cite{alain2015variance,gopal2016adaptive}. Akin to CL, HEM also has various applications, and the famous boosting algorithm~\cite{freund1996experiments} in ensemble learning also takes the same strategy by upweighting the wrongly-classified examples. 

So which strategy should we apply to our own scenario, ``easier first'' as CL or ``harder first'' as HEM? It remains an unsolved problem to be carefully considered. Theoretically, under different settings, both CL and HEM strategies can benefit the learning as long as the ``curriculum'' is positively correlated with the optimal utility\footnote{The optimal utility is $\sum_{i \in \mathcal{D}} e^{-l_i (\theta^*)}$, where $\mathcal{D}$ is the training dataset, $l_i (\theta^*)$ is the loss on the $i$-th example calculated by the optimal model $\theta^*$.}~\cite{hacohen2019power}. However, this criterion is very hard to verify. More intuitively, Chang et al.~\cite{chang2017active} point out that CL is more suitable for the scenarios with more noisy labels or outliers to improve the model robustness and convergence rate, while HEM is more beneficial for cleaner datasets and leads to faster and more stable SGD. One should also note that if the target task is very difficult, CL will be more preferable to HEM, since CL is able to result in a more effective training process through the easier/smoother versions.

\begin{figure*}[htbp]
    \centering
    \centerline{\includegraphics[width=0.85\linewidth]{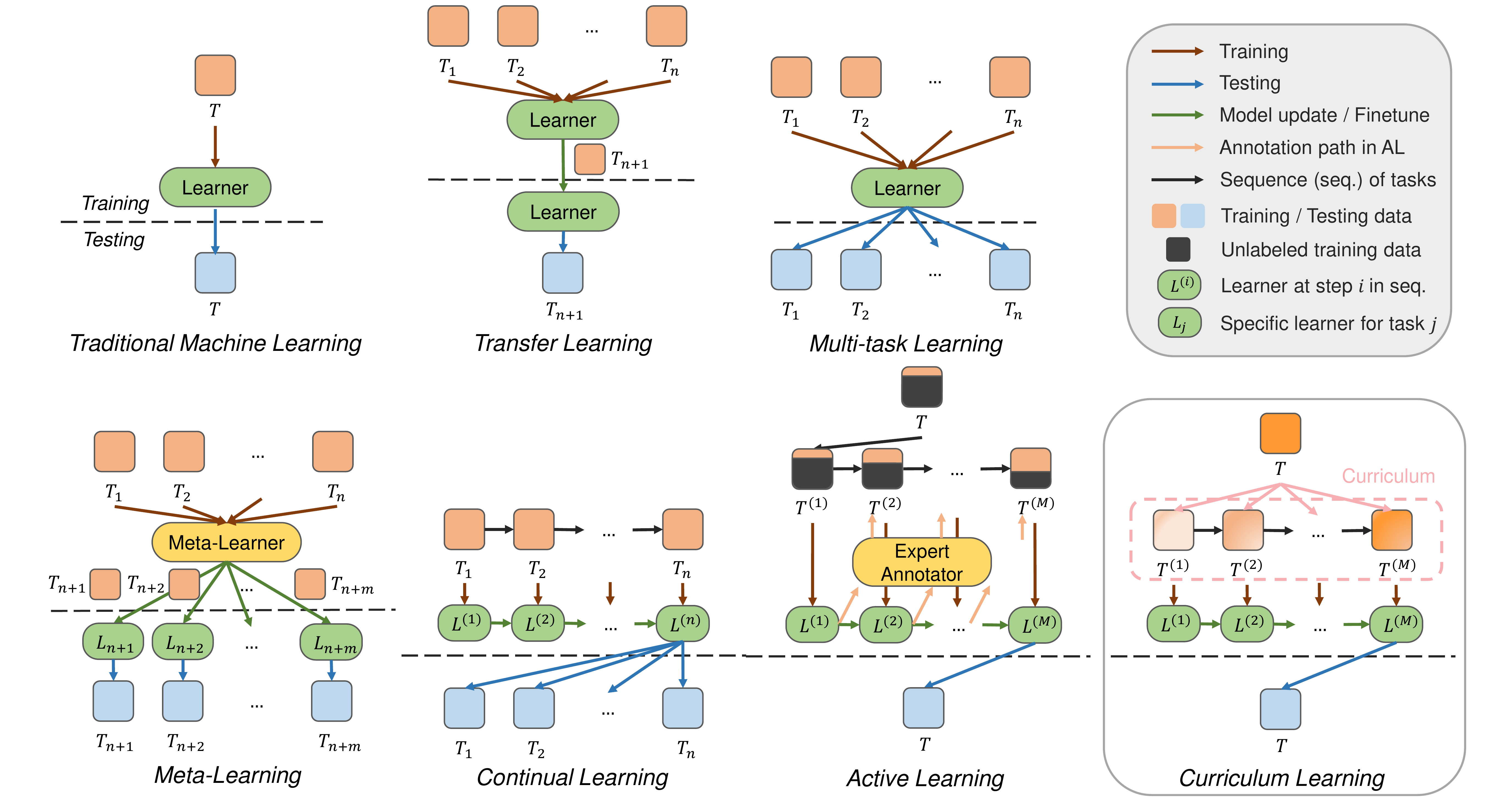}}
    \caption{Illustration of different machine learning paradigms from the perspective of data distribution. Different paradigms aim to solve different distribution discrepancies among training and testing data, while we see similar mechanisms among some of them, which help us understand their connections and may potentially inspire new methodologies. For curriculum learning, we illustrate Definition 2, and the curriculum can be both predefined and automatically learned. Note that $T_j$ stands for different tasks, while $T^{(i)}$ is the modified distribution at the $i$-th step in training.}
    \label{fig:x_learning}
\end{figure*}

An alternative is to combine the two strategies together with a trade-off policy. For example, Pi et al.~\cite{pi2016self} embed the self-paced regularizers into the objective of boosting algorithm, which simultaneously enhances the learning effectiveness (by boosting) and robustness (by SPL).
Besides, Chang et al.~\cite{chang2017active} propose to select the most uncertain examples according to the prediction history, which is consistent with the variance reduction strategies in active learning~\cite{settles2009active}. The uncertain examples are predicted both incorrectly and correctly in history and are thus neither too easy (always correct) nor too difficult (always incorrect). It is worth mentioning that the fully automatic CL methods (e.g., RL Teacher in Sec.~\ref{subsubsec:RL-teacher}) would also be an ideal choice when it is hard to choose between ``easier first'' CL and ``harder first'' HEM.

From a higher perspective, both the original CL (Definition 1) and HEM belong to the instance selection or example reweighting, defined as data-level generalized CL (Definition 2) in Sec.~\ref{sec:2-definition}. As argued in~\cite{ren2018learning}, one crucial advantage of reweighting examples is robustness against training set biases. The biases include class imbalance and label noise, both of which have been studied as typical problems of machine learning with various practical methods (e.g., ~\cite{chawla2002smote,khan2017cost} for the former and~\cite{natarajan2013learning,reed2014training,goldberger2016training,li2017learning} for the latter). By reweighting examples, HEM prioritizes higher-loss examples which more likely belong to minority classes, and thus alleviates class imbalance bias. On the other hand, CL favors lower-loss examples which are more likely to be clean data, and thus reduces the label noise bias. When assumptions on the training set biases are uncertain, many fully automatic CL methods are designed to reweight the examples to achieve a certain goal of learning, e.g., training efficiency~\cite{graves2017automated,matiisen2019teacher}, valid set accuracy~\cite{ren2018learning,wang2020optimizing,kumar2019reinforcement}, etc.

\vspace{-2mm}
\subsection{Relationship between CL and Other Concepts}
\label{subsec:5-2-relation-graph}

From the perspective of data distribution, different machine learning paradigms focus on different settings on data distribution discrepancy, which is illustrated in Fig.~\ref{fig:x_learning}. For example, transfer learning~\cite{pan2009survey} aims at alleviating the discrepancy between source tasks $\{T_i\}_{i=1}^n$ and target task by transferring through model parameters of the learner. Meta-learning~\cite{hospedales2020meta} mitigates the discrepancy between multiple source tasks $\{T_i\}_{i=1}^n$ and target tasks $\{T_i\}_{i=n+1}^{n+m}$ by learning common meta-knowledge on learning algorithms across tasks. Continual learning~\cite{delange2021continual} eases the discrepancy among an online sequence of tasks by updating one learner to defy forgetting. From this view, Data-level Generalized Curriculum Learning (Definition 2) smooths the discrepancy between the testing distribution and training distribution by a sequence of reweighting, which results in a gradual optimization process towards the target.

With Fig.~\ref{fig:x_learning}, we can see the differences and connections between CL and other concepts, which may inspire new ideas.
(i) \textbf{CL v.s. Transfer Learning (TL):} as pointed out by Bengio et al.~\cite{bengio2009curriculum}, CL can be seen as a special form of TL where the initial tasks are used to guide the learner so that it will perform better on the final task. Thus, CL is naturally suitable for TL settings like domain adaption~\cite{shu2019transferable,zhang2019curriculum}. The green arrows also show that CL is a sequence of TL throughout the curriculum. 
(ii) \textbf{CL v.s. Multi-task Learning (MTL):} we can regard the $T$ in CL as a distribution of tasks and the $n$ tasks in MTL are sampled from this distribution. CL then provides a sequence of task distributions to guide MTL, which is empirically proven helpful~\cite{li2016multitask,pentina2015curriculum,sarafianos2017curriculum,graves2017automated}.
(iii) \textbf{CL v.s. Meta-Learning (ML):} although ML and CL seem quite different in Fig.~\ref{fig:x_learning}, we argue that ML is highly related to automatic CL (AutoCL). In fact, the teaching policy (i.e., curriculum) in AutoCL can be regarded as the meta-knowledge in ML to optimize the student's progress~\cite{hospedales2020meta}, from which view AutoCL is a specific form of CL. In essence, ML is about learning to learn and AutoCL is about learning to teach~\cite{fan2018learning}, i.e., they both aim to optimize the hyperparameters of algorithms from different views of students and teachers. Therefore, it is no wonder that ML is shown effective for AutoCL designs~\cite{ren2018learning,wang2020optimizing,shu2020meta}, and shall inspire more AutoCL ideas. We also advocate the integration of ML and AutoCL to enable fully automatic machine learning and teaching. 
(iv) \textbf{CL v.s. Continual Learning (ContL):} although both of them involve a sequence of tasks, the settings are quite different. Specifically, with a different distribution, the tasks $\{T_i\}_{i=1}^n$ in ContL are predefined and fixed. While in CL, derived from the same distribution $T$, the distributions $\{T^{(i)}\}_{i=1}^M$ in $M$ steps can be flexibly adjusted by the curriculum. However, we argue that within each task in ContL, CL methods may help to improve robustness and defy forgetting by the transfer between preceding tasks and the current task. 
(v) \textbf{CL v.s. Active Learning (AL)}: AL~\cite{settles2009active} is the most analogous paradigm to CL in Fig.~\ref{fig:x_learning}, both of which involves dynamic data selection. In AL, an active learner achieves great performance with fewer labeled data via generating queries to ask an expert to annotate several unlabeled instances for further training. The goals of CL and AL are different: the former improves performance and accelerates convergence in supervised, weakly-supervised, and unsupervised settings, while the latter is designed for label-saving training in the semi-supervised setting. However, the criteria for data selection can somehow be shared among CL and AL, and recent works~\cite{lin2017active,tang2019self} have made efforts to combine SPL with AL to utilize the complementariness between the criteria.

\section{Future Directions of CL}
\label{sec:6-future-directions}

We conclude this paper with some ongoing or future directions of CL, which are worthy of discussion:



\textbf{Evaluation benchmarks.} 
Although various CL methods have been proposed and demonstrated effective, few works have made efforts on evaluating them with general benchmarks. 
In existing literature, the datasets and metrics are diverse in different applications. For instance, the CIFAR datasets with different label corruption settings are widely used to evaluate CL methods on image classification with accuracy metric~\cite{jiang2018mentornet,saxena2019data,wang2020optimizing} , and the WMT datasets are widely chosen to evaluate CL methods for neural machine translation with BLEU metric~\cite{platanios2019competence,kumar2019reinforcement,liu2020norm}. However, it is challenging to design a unified dataset with unified metrics to evaluate and compare the CL algorithms.
Such a benchmark may incorporate datasets for different applications (e.g., CV, NLP, recommendation, etc.) with different noise levels (e.g., clean, weakly-supervised, etc.). Accordingly, evaluation metrics on the relative performance boost, convergence speedup, additional computational cost, etc., should also be carefully designed.
The challenges are three-fold: (i) Dataset construction: the data of different applications have different levels of sparsity, heterogeneity, noisiness, etc. (ii) Metric design: different applications naturally need different metrics, and their urgency of requirements for convergence speed is also different. (iii) Ground-truth curriculum: most CL literature does not provide an oracle curriculum to evaluate whether the algorithm-based curriculum is reasonable. Therefore, it would be interesting to design such an ideal curriculum in the benchmark to compare CL methods more intuitively.

\textbf{More advanced theories.} Existing theoretical analyses in Sec.~\ref{subsec:3-1-theoretical-analysis} provide different angles for understanding CL. Nevertheless, more theories are still required to help us reveal why typical CL (Definition 2 in Sec.~\ref{sec:2-definition}) is effective. For example, if the dataset has no noise, are there any bounds for the effectiveness of CL? What is the actual effect of each condition in Definition 2, i.e., increasing dataset size/variance and increasing difficulty? Besides, the fully automatic CL methods in Sec.~\ref{subsubsec:RL-teacher} and ~\ref{subsubsec:other-automatic-CL} also need more theoretical guarantees on their effectiveness. Moreover, a remaining fundamental question is to theoretically reveal the relations between the data distribution, task objective, and the best training strategy among ``easier first'' (CL), ``harder first'' (HEM), and other strategies. Theories on this topic shall provide the basis for the application of CL in a specific task.

\textbf{More CL algorithms and various applications.} Automatic CL (Sec.~\ref{subsec:4-3-automatic-CL}) provides the potential application values for CL in wider research areas and has become a cutting-edge direction. Therefore, one promising direction is to design more automatic CL methodologies with different optimizations (e.g., bandit algorithms, meta-learning, hyperparameter optimization, etc.) and different objectives (e.g., data selection/reweighting, finding the best loss function or hypothesis space, etc.). Moreover, as shown in ~\cite{pi2016self,lin2017active,tang2019self}, CL methods can be incorporated with other strategies like boosting and AL to achieve improvement. 
In addition to methodologies, more efforts should be made to explore the power of CL in more various applications, including both cutting-edge research areas (e.g., meta-learning, continual learning, NAS, graph neural network, self-supervised learning, etc.) and traditional machine learning topics (e.g., clustering, regression, etc.).
Although the directions mentioned above may adopt Definition 3 of CL as a sequence of training criteria in Sec.~\ref{sec:2-definition}, the spirit of imitating the human curriculum shall drive more breakthroughs in the machine learning community.

\begin{IEEEbiography}[{\includegraphics[width=1in,height=1.25in,clip,keepaspectratio]{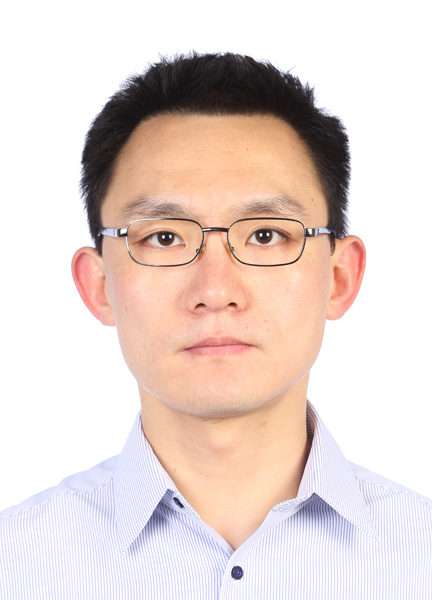}}]{Xin Wang}
 is currently an Assistant Professor at the Department of Computer Science and Technology, Tsinghua University. He got both of his Ph.D. and B.E degrees in Computer Science and Technology from Zhejiang University, China. He also holds a Ph.D. degree in Computing Science from Simon Fraser University, Canada. His research interests include relational media big data analysis, multimedia intelligence and recommendation in social media. He has published several high-quality research papers in top conferences including ICML, KDD, WWW, SIGIR ACM Multimedia etc. He is the recipient of 2017 China Postdoctoral innovative talents supporting program. He receives the ACM China Rising Star Award in 2020. 
\end{IEEEbiography}

\begin{IEEEbiography}[{\includegraphics[width=1in,height=1.25in,clip,keepaspectratio]{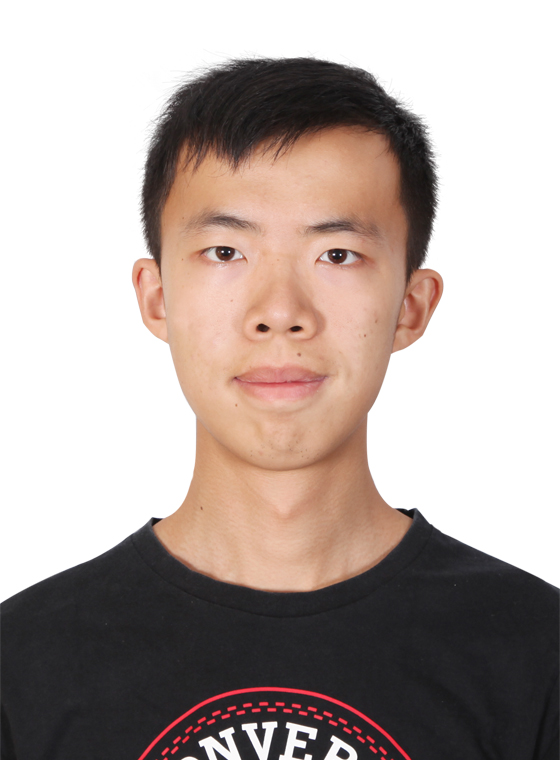}}]{Yudong Chen} is a graduate student at the Department of Computer Science and Technology, Tsinghua University. His research interests include machine learning, data mining, and multimedia analysis.
\end{IEEEbiography}

\begin{IEEEbiography}[{\includegraphics[width=1in,height=1.25in,clip,keepaspectratio]{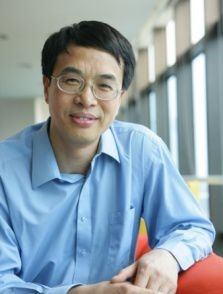}}]{Wenwu Zhu}
 is currently a Professor and the Vice Chair of the Department of Computer Science and Technology at Tsinghua University. 
 His research interests are in the area of data-driven multimedia networking and Cross-media big data computing. He has published over 350 referred papers and is the inventor or co-inventor of over 50 patents. He received eight Best Paper Awards, including ACM Multimedia 2012 and IEEE Transactions on Circuits and Systems for Video Technology in 2001 and 2019.  
 
 He served as EiC for IEEE Transactions on Multimedia from 2017-2019. He served in the steering committee for IEEE Transactions on Multimedia (2015-2016) and IEEE Transactions on Mobile Computing (2007-2010), respectively.
 He is an AAAS Fellow, IEEE Fellow, SPIE Fellow, and a member of The Academy of Europe (Academia Europaea).
\end{IEEEbiography}




%








\end{document}